\documentclass{article}
\usepackage{fullpage}

\usepackage[utf8]{inputenc} 
\usepackage[T1]{fontenc}    
\usepackage{url}            
\usepackage{booktabs}       
\usepackage{amsfonts}       
\usepackage{nicefrac}       
\usepackage{microtype}      
\usepackage{soul}
\usepackage{graphicx}
\usepackage{amsmath}
\usepackage{outlines}
\usepackage{xurl}

\usepackage[colorlinks=true,
citecolor=blue,
filecolor=black,
linkcolor=blue,
urlcolor=blue]{hyperref}

\usepackage{amsmath,amsfonts,bm}

\usepackage{xcolor}
\usepackage{colortbl}

\def\eqref#1{equation~\ref{#1}}

\def\rvx{{\mathbf{x}}}

\def\rmW{{\mathbf{W}}}

\def\mW{{\bm{W}}}
\def\mX{{\bm{X}}}
\def\mY{{\bm{Y}}}

\DeclareMathAlphabet{\mathsfit}{\encodingdefault}{\sfdefault}{m}{sl}
\SetMathAlphabet{\mathsfit}{bold}{\encodingdefault}{\sfdefault}{bx}{n}

\def\sR{{\mathbb{R}}}

\usepackage{multirow}
\usepackage{paralist}

\usepackage{multicol}
\usepackage{diagbox}

\usepackage[ruled,noend]{algorithm2e}

\SetCommentSty{mycommfont}

\usepackage{here}

\usepackage{amsmath,amssymb,amsfonts,amsbsy,amsfonts,latexsym}
\usepackage{makecell}
\usepackage{xcolor}
\usepackage{colortbl}

\usepackage{tabularx,colortbl,xcolor}
\usepackage[normalem]{ulem}
\useunder{\uline}{\ul}{}

\usepackage{enumitem}

\usepackage{xparse}

\SetKwInput{KwInput}{Input}
\SetKwInput{KwRequire}{Require}

\usepackage{longtable}

\NewDocumentCommand{\var}{O{s} m O{}}{%
  \ensuremath{#1_{#2}^{#3}}
}
\usepackage{siunitx}

\newcommand{\commentout}[1]{}

\definecolor{light-gray}{gray}{0.80}

\newcommand\appref{Appendix~\ref}

\newcommand\fref{Figure~\ref}
\newcommand\tref{Table~\ref}
\newcommand\sref{Section~\ref}

\usepackage{amsthm}

\usepackage{amsthm}

\newtheorem{mytheorem}{Theorem}
\newtheorem{assumption}[mytheorem]{Assumption}
\newtheorem{lemma}[mytheorem]{Lemma}
\newtheorem{remk}[mytheorem]{Remark}

\usepackage{subfig}

\def\loss{\mathcal{L}}

\usepackage{pifont}
\newcommand{\cmark}{\ding{51}}%
\newcommand{\xmark}{\ding{55}}%

\newcommand{\qat}{QAT\xspace}
\newcommand{\ptq}{PTQ\xspace}
\newcommand{\OURS}{ZeroQuant\xspace}
\newcommand{\OURSlwd}{ZeroQuant-LKD\xspace}
\newcommand{\mhsa}{MHSA\xspace}
\newcommand{\ffc}{FFC\xspace}
\newcommand{\kd}{KD\xspace}
\newcommand{\lwd}{LKD\xspace}
\newcommand{\wikitext}{Wikitext-2\xspace}
\newcommand{\wikitextone}{Wikitext-103\xspace}
\newcommand{\bert}{BERT\xspace}
\newcommand{\bertbase}{BERT$_\text{{base}}$\xspace}
\newcommand{\bertlarge}{BERT$_{\text{large}}$\xspace}
\newcommand{\gpt}{GPT-3\xspace}
\newcommand{\gpttf}{GPT-3$_{\text{350M}}$\xspace}
\newcommand{\gptot}{GPT-3$_{\text{1.3B}}$\xspace}

\newcommand{\gptneox}{GPT-NeoX$_{\text{20B}}$\xspace}
\newcommand{\gptj}{GPT-J$_{\text{6B}}$\xspace}

\usepackage{chngcntr}
\usepackage{adjustbox}

\begin{document}

\title{\OURS: Efficient and Affordable Post-Training Quantization for Large-Scale Transformers}

\author{Zhewei Yao\thanks{Code will be released soon as a part of \url{https://github.com/microsoft/DeepSpeed}}~, Reza Yazdani Aminabadi, Minjia Zhang\\ Xiaoxia Wu, Conglong Li, Yuxiong He  \vspace{0.2cm} \\  Microsoft \\ {\tt \small\{zheweiyao, yazdani.reza, minjiaz, xiaoxiawu, conglong.li, yuxhe\}@microsoft.com}
}

\date{}
\maketitle

\begin{abstract}
How to efficiently serve ever-larger trained natural language models in practice has become exceptionally challenging even for powerful cloud servers due to their prohibitive memory/computation requirements.
In this work, we present an efficient and affordable post-training quantization approach to compress large Transformer-based models, termed as \OURS. 
\OURS is an end-to-end quantization and inference pipeline with three main components: 
(1) a fine-grained hardware-friendly quantization scheme for both weight and activations; 
(2) a novel affordable layer-by-layer knowledge distillation algorithm (\lwd) even without the access to the original training data;
(3) a highly-optimized quantization system backend support to remove the quantization/dequantization overhead.
As such, we are able to show that:
(1) \OURS can reduce the precision for weights and activations to INT8 in a cost-free way for both \bert and \gpt-style models with minimal accuracy impact, which leads to up to 5.19x/4.16x speedup on those models compared to FP16 inference;
(2) \OURS plus \lwd affordably quantize the weights in the fully-connected module to INT4 along with INT8 weights in the attention module and INT8 activations, resulting in 3x memory footprint reduction compared to the FP16 model;
(3) \OURS can be directly applied to two of the largest open-sourced language models, including \gptj and \gptneox, for which our INT8 model achieves similar accuracy as the FP16 model but achieves up to 5.2x better efficiency.
\end{abstract}
\section{Introduction}
\label{sec:intro}

Large-scale natural language models have been widely adopted in different applications, e.g., natural language understanding using BERT~\cite{tenney2019bert} and generation tasks using GPT-style models~\cite{radford2019gpt}. 
Although those models have achieved cutting-edge accuracy results, as the model size keeps increasing dramatically, the requirements of memory footprint and the computational cost to deploy them become a major bottleneck, even on cloud servers with powerful GPU devices. 

One promising way to alleviate this challenge is quantization, which can reduce the bit precision for both weight and activations for lower memory footprint and faster compute (e.g., INT8 Tensor cores on T4/A100). 
However, quantization usually requires retraining (also known as quantization aware training, or \qat in short) to recover the accuracy degradation from representation loss of weight and activations. 
To enable \qat, the full training pipeline is usually required, including the training data and compute resources, to finetune the model.
Access to those components is now oftentimes not available, and \qat is also a time-consuming process, particularly for those large-scale models.

Recently, zero-shot quantization~\cite{cai2020zeroq,nagel2019data} and post-training quantization (\ptq)~\cite{nagel2020up,liu2021post} are proposed to address the training-data access and compute  requirement challenges since \ptq generally requires no (or minimal) retraining. 
But most of those works primarily focus on computer vision problems on relatively small scales.  
More recently, \cite{bondarenko2021understanding} shows promising \ptq results on BERT. 
However, (1) its main focus is on high-precision quantization (INT8/FP16) on \bertbase, (2) it does not consider other billion-scale generative models (\gpt-style models~\cite{brown2020language}). More importantly, most of these works do not report real latency improvement, putting the usefulness of these methods in improving inference latency into question.
For example, existing work often do not discuss the quantization/dequantization cost associated with different quantization schemes, which in fact has a big impact to the performance benefit of using low precision. 

Besides, for extreme quantization (e.g., INT4), knowledge distillation is usually used to boost performance, which adds another source of expensive computation cost as compared to \qat. 
Furthermore, in order to achieve better accuracy performance, hidden-states knowledge distillation, e.g.,~\cite{bai2020binarybert,zhang2020ternarybert}, is usually applied for the quantized model. 
This would put significant pressure on the GPU memory and the compute resource requirement since both the teacher and student models needed to be loaded into the GPU memory for training.

In this paper, we present \OURS, an end-to-end post-training quantization and inference pipeline, to address those challenges, targeting both INT8 and INT4/INT8 mixed-precision quantization. 
Specifically, our contributions are:
\begin{itemize}[noitemsep, nolistsep, labelindent=0pt, leftmargin=*]
    \item We apply fine-grained hardware-friendly quantization schemes on both weight and activations, i.e., group-wise quantization for weight and token-wise quantization for activations. 
    Both quantization schemes can significantly reduce the quantization error and retain hardware acceleration properties.
    \item We propose a novel layer-by-layer knowledge distillation method (\lwd) for INT4/INT8 mixed-precision quantization, where the neural network is quantized layer-by-layer through distillation with minimal iterations and even without the access to the original training data. 
    As such, at any given moment, the device memory is primarily populated only with a single extra layer's footprint, making billion-scale model distillation feasible with limited training budget and GPU devices.
    \item We develop a highly optimized inference backend, which eliminates the expensive computation cost of quantization/dequantization operators, enabling latency speedups on INT8 Tensor cores on modern GPU hardware. 
    \item Our empirical results show that:
        \begin{itemize}[noitemsep, nolistsep, labelindent=0pt, leftmargin=*]
            \item \OURS enables quantizing \bert and \gpt-style models into INT8 weight and activations to retain accuracy without incurring any retraining cost. 
            Compared to FP16 inference, our INT8 model achieves up to 5.19x/4.16x speedup on \bertbase/\gpttf on A100 GPUs.
            \item \OURS plus \lwd can do INT4/INT8 mixed-precision quantization for \bert and \gpt-style models.
            This results in a 3x memory footprint reduction with marginal accuracy loss as compared to the FP16 model.
            Also, thanks to the lightweight of \lwd, 
            we can finish the quantization process in 33s (10 minutes) for \bertbase (\bertlarge). 
            We also demonstrate that \lwd can use other datasets to achieve similar performance to the original training data.
            \item We demonstrate the scalability of \OURS on two of the largest open-sourced language models, i.e, \gptj and \gptneox, with INT8 quantization. 
            \OURS can achieve 3.67x speedup over the FP16 model for \gptj and (2) reduce the GPU requirement for inference from 2 to 1 and latency from 65ms to 25ms for \gptneox (i.e., 5.2x better system efficiency in total).
        \end{itemize}
\end{itemize}

\section{Related Work}
\label{sec:related_work}

Model compression has been explored from different aspects~\cite{han2015learning, li2016pruning,mao2017exploring,lecun1990optimal,michel2019sixteen, fan2019reducing, gordon2020compressing, raganato2020fixed,dong2019hawq,yao2021mlpruning, mao2020ladabert,hinton2015distilling,sanh2019distilbert, sun2019patient,jiao2019tinybert, sun2020mobilebert, wang2020minilm, lan2019albert, dehghani2018universal,liu2021post,kim2021bert}. 
Among those, quantization is one of the most promising directions as it directly reduces the memory footprint and compute intensity. 
Here, we focus on quantization for NLP models and briefly discuss the related work.

The majority of quantization works can be categorized into quantization-aware training (\qat).
\cite{shen2020q,zafrir2019q8bert} are the first few works to quantize BERT models using integer numbers for both weight and activations. 
Particularly, \cite{shen2020q} utilizes Hessian information to push the weight bit-precision to even INT2/INT4, and it also proposes group-wise quantization to quantize the weight matrix in a more fine-grained granularity compared to single matrix quantization.
\cite{fan2020training} introduces quantization noise to alleviate the variations of \qat. 
\cite{zhang2020ternarybert,bai2020binarybert} leverage very expensive knowledge distillation~\cite{hinton2015distilling} and data augmentation~\cite{jiao2019tinybert} to ternarize/binarize weights. 
\cite{jin2021kdlsq} combines knowledge distillation~\cite{jiao2019tinybert} and learned step size quantization~\cite{esser2019learned} to quantize the weight to 2--8 bits. 
Recently, \cite{tao2022compression} also uses knowledge distillation to compress GPT-2 models on task-specific problems to INT2. 
All those works quantize models using the original training datasets.
More importantly they need retraining or finetuning the full model to recover the accuracy, and such compute cost on extra-large models, like ~\cite{smith2022using,chowdhery2022palm}, can be hardly affordable for most research labs or practitioners. 

One solution to overcome the compute cost challenge is post-training quantization (\ptq). However, \ptq often induces a significant drop in accuracy because the network can be sensitive to quantization errors.
Along this line, one of the first works applied to Transformer-based~\cite{vaswani2017attention} models is~\cite{zadeh2020gobo}. 
The authors introduce centroid-based quantization method, where outlier numbers use FP32 format and the rest numbers are quantized using non-uniform quantization. 
As such, it is hard to get the real inference latency benefit on general compute accelerators, e.g., CPU and GPU, because the parallel processing units in these hardware do not support efficient computation of mixed data types. 
More recently, \cite{bondarenko2021understanding} introduces high-precision activation quantization (FP16) for part of the model to overcome the high dynamic activation ranges. 
However, to the best of our knowledge, 
(1) How to apply \ptq on \gpt-style models while achieving high accuracy has not been studied in any of previous work yet; 
(2) How to apply \ptq on billion (or even a dozen of billions) scale model is still under-explored; 
(3) Efficient inference system backend is still missing, especially for fine-grained quantization schemes, making it hard to achieve low latency on commodity hardware. 
\OURS resolves all those limitations by considering the system backend into the algorithm design and we verify its capability on both \bert and large-scale \gpt-style (up to 20 billion, i.e., \gptneox) models for various tasks.  
\section{Background and Challenge}
\label{sec:ptq_challenge}

We give a brief overview of the transformer architecture and quantization background in~\appref{sec:background}. 
Please refer to~\cite{vaswani2017attention} and~\cite{gholami2021survey} for more details about the transformer architecture and quantization.

Post-training quantization (\ptq) exhibits great compression efficiency compared to quantization-aware training (\qat) since \ptq is usually applied to quantize the model without retraining. 
A common strategy of \ptq is to feed the training data to the network and calibrate the scaling factor, $S$, using the running mean. 
Please see~\appref{sec:ptq_calibrate} for more details.

Some work has been done for \bertbase models~\cite{bondarenko2021understanding} with INT8 weight and mixed INT8/FP16 activation quantization. 
However, there is no investigation for (1) even lower bit-precision \ptq on \bert models and (2) large-scale \gpt-style models. 
Here, we briefly discuss the challenge of the application of \ptq on both \bert (in~\appref{sec:ptq_challenge_of_bert}) and \gpt-style models.

\begin{figure}[h]
\centering
\includegraphics[width=0.45\linewidth]{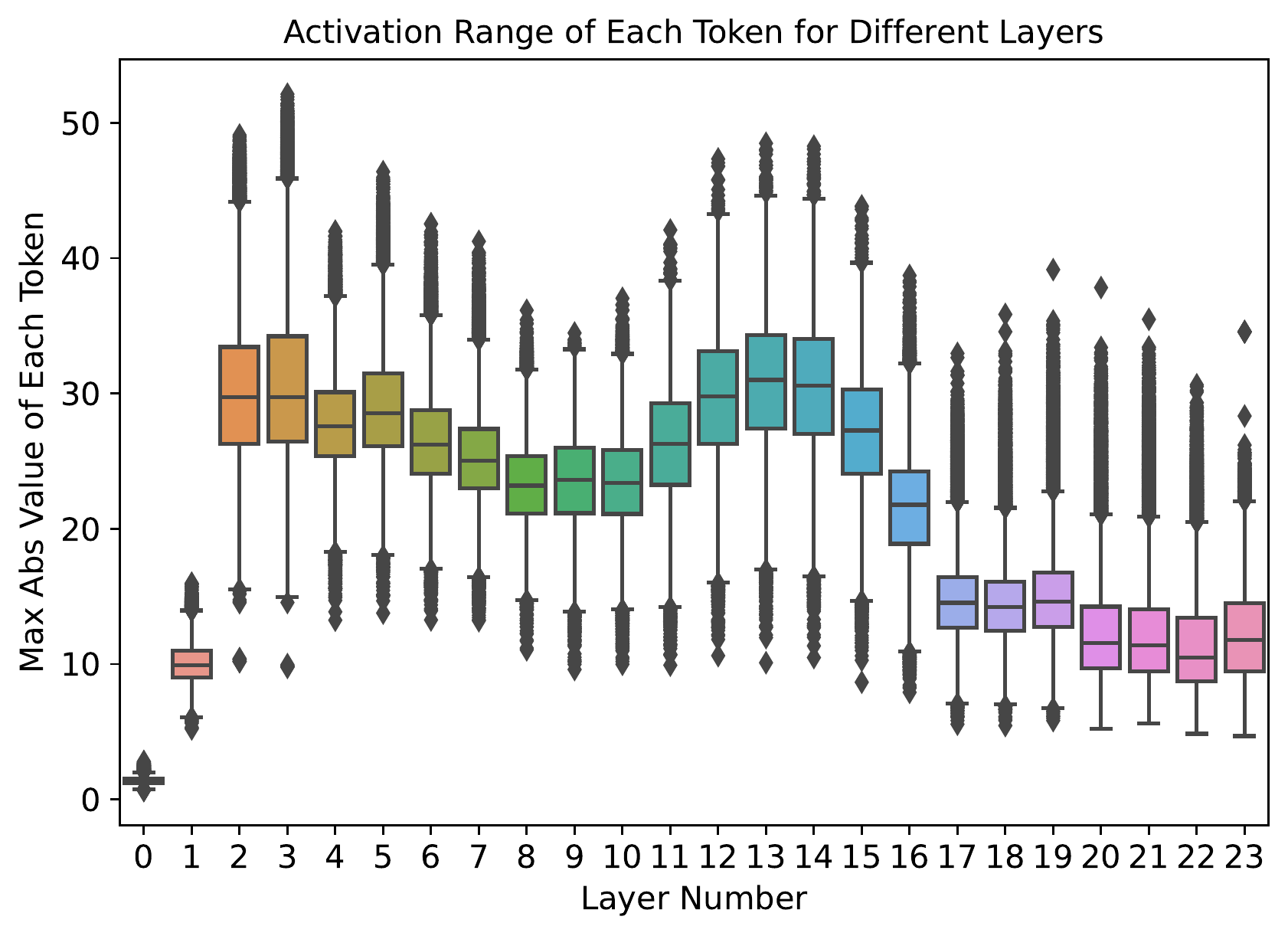}
\includegraphics[width=0.45\linewidth]{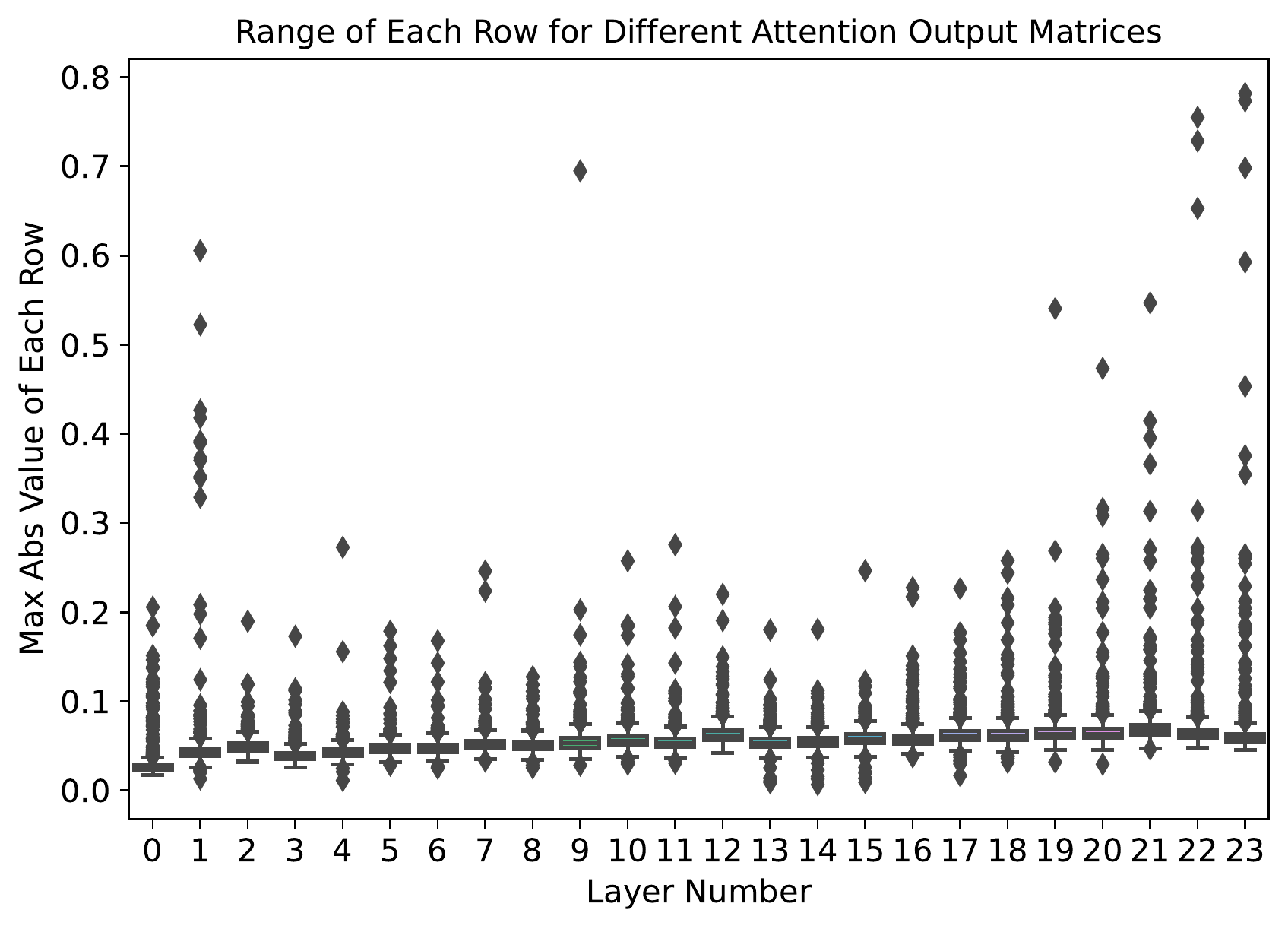}
\caption{
The activation range (left) and row-wise weight range of the attention output matrix  (right) of different layers on the pretrained \gpttf. 
See~\fref{fig:activation_weight_range} for the results of \bertbase.
}
\label{fig:activation_weight_range_gpt}
\end{figure}

The results of \gpttf with \ptq are shown in~\tref{tab:gpt3_investigation}.
As can be seen, the INT8 activation quantization (i.e., the row of W16A8) causes the primary accuracy loss. 
Further pushing the weight to INT8 (i.e., the row of W8A8) does not change the accuracy of zero-shot evaluation tasks but leads the causal language modeling task (\wikitext) to worse perplexity score, 
which demonstrates the sensitivity of generation tasks as compared to other zero-shot evaluation problems. 
For W4/8A16, on some accuracy-based tasks, \gpttf still achieves reasonable performance like OpenBookQA but it loses accuracy on the majority of the rest tasks. 
Particularly, for \wikitext, \gpttf with W4/8A16 cannot generate any meaningful text anymore. 
Please also see~\appref{sec:ptq_challenge_of_bert} for the analysis for \bert.

\begin{table}[t]
\caption{
Post training quantization results of \gpttf on 20 zero-shot evaluation datesets.
Here WxAy means x-/y-bit for weight/activation.
Particularly, for W4/8, we quantize the \mhsa's weight to INT8 and \ffc's weight to INT4.
Please see~\tref{tab:gpt3_350_full_table} for the results of all 20 tasks.
}\centering
\label{tab:gpt3_investigation}
\begin{adjustbox}{width=0.9\linewidth}
\centering
\begin{tabular}{lcccccccccccccc }
\toprule
Precision     & Lambada ($\uparrow$)    & PIQA ($\uparrow$) & OpenBookQA ($\uparrow$) & RTE ($\uparrow$) & ReCoRd ($\uparrow$) & Ave. 19 Tasks ($\uparrow$) & \wikitext ($\downarrow$) \\
\midrule
W16A16     & 49.3 & 66.3 & 29.4 & 53.8 & 75.1 & 38.9 & 21.5  \\
W8A16      & 49.3 & 66.1 & 29.6 & 54.2 & 74.8 & 38.5 & 22.1  \\
W16A8      & 44.7 & 64.8 & 28.2 & 52.7 & 69.2 & 37.8 & 24.6  \\
W8A8       & 42.6 & 64.1 & 28.0 & 53.1 & 67.5 & 37.8 & 26.2  \\
W4/8A16    & 0.00 & 51.4 & 30.2 & 52.7 & 16.1 & 28.9 & 1.76e5\\
\bottomrule
\end{tabular}
\end{adjustbox}
\end{table}

\noindent
\textbf{Dynamic Activation Range}
To investigate why INT8 activation leads to significant accuracy drop for both \bert and \gpt-style models, we plot the token-wise (i.e., the hidden state of each token) range of each activation for different transformer layers of \gpttf in~\fref{fig:activation_weight_range_gpt} (left).
As can be seen, different tokens have dramatically different activation ranges. 
For example, the maximum range of the last layer is around 35 but the minimum range is close to 8. This larger variance in the activation range makes it difficult to use a fixed quantization range (usually the maximum value) for all tokens to retain the prediction accuracy, because the limited representation power for small range tokens is going to hurt the accuracy performance. 

\noindent
\textbf{Different Ranges of Neurons in Weight Matrices}
Similarly, we plot the row-wise (i.e., the output dimension) weight range of the attention output matrix ($\mW_{o}$) of \gpttf in~\fref{fig:activation_weight_range_gpt} (right). 
There is a 10x difference between the largest magnitudes of different rows and this leads to the worse generation performance of the INT8 weight \ptq. 
This also makes it very challenging when INT4 quantization is applied as the INT4 only has 16 numbers and a 10x smaller range leads to 2 (or 3) numbers for the representations of those smaller-range rows.

This analysis results also indicate why more expensive hidden-states knowledge distillation~\cite{bai2020binarybert,li2016ternary} is used for ultra-low precision quantization to close the accuracy gap. 
However, as the training cost of knowledge distillation for large-scale models is too high, a lightweight and efficient method is desirable for \ptq. 
\section{Methodology}
\label{sec:methodology}
\subsection{Fine-grained Hardware-friendly Quantization Scheme}

As shown in~\sref{sec:ptq_challenge}, even applying INT8 \ptq to \bert/\gpt-style models leads to significant accuracy degradation. 
The key challenge is the representation of INT8 cannot fully capture the different numerical ranges of different rows in weight matrices and different activation tokens. 
One way to address this is to use group-wise (token-wise) quantization for the weight matrix (activations).

\noindent
\textbf{Group-wise Quantization for Weights} Group-wise weight matrix quantization has first been proposed in~\cite{shen2020q}, 
where a weight matrix $\rmW \in \sR^{n\times m}$ is partitioned in to $g$ groups, and each group is quantized separately. 
However, in~\cite{shen2020q}, the authors only apply this for quantization aware training.
More importantly, they do not consider the hardware efficiency constraint and they do not have a system backend support.
As such, they lack the real latency reduction benefit.

In our design, we consider the hardware constraint from Ampere Architecture of GPUs (e.g, A100), where the compute unit is based on Warp Matrix Multiply and Accumulate (WMMA) tiling size~\cite{nvidia_wmma} to achieve the best speedup.
Later, we will show that our group-wise quantization leads to much better accuracy as compared to single-matrix quantization due to its finer-granularity quantization while still achieving great latency reduction.

\noindent
\textbf{Token-wise Quantization for Activations}
As mentioned in~\sref{sec:ptq_challenge} and \appref{sec:quantization_background}, a common practice for existing \ptq work is to use static quantization for activation, where the min/max range is calculated at an offline calibration phase. 
Such a method might be sufficient for small scale models where the variance in the activation range is small.
However, as analyzed in~\sref{sec:ptq_challenge}, there is a huge variance in the activation range for large-scale transformer models such as \gpttf and \bertbase. 
As such, a static quantization scheme (often applied to all tokens/samples) would lead to significant accuracy drop. 
One natural idea to overcome this issue is to adopt finer-grained token-wise quantization and dynamically calculate the min/max range for each token to reduce the quantization error from activations. 
Our evaluation in ~\sref{sec:results} also shows that token-wise quantization for activation significantly improves the accuracy of \gpt-style and \bert models. 

However, directly applying token-wise quantization using existing DL frameworks, such as the PyTorch quantization suite, would lead to significant quantization and dequantization cost because token-wise quantization introduces additional operations that lead to expensive data movement overhead between the GPU compute units and the main memory. 
To address this issue, we build a highly optimized inference backend for token-wise quantization of transformer models. For example, the inference backend of \OURS employs so called \emph{kernel fusion} technique to fuse quantization operator with its previous operator, like layer normalization, to alleviate the data movement cost from token-wise quantization. Similarly, the dequantization cost of the different GeMMs' output is alleviated by scaling the INT32 accumulation using both 
the weight and activation quantization scales, before writing the final FP16 result back to the main memory for the next FP16 operator (like GeLU). 
Those optimization will be discussed in more details in~\sref{sec:system_optimization}.

Token-wise quantization can significantly reduce the representation error for quantized activations.
Also, as it does not need to calibrate the activation range, later we will show that there is no quantization-related cost (e.g., activation range calibration) for a moderate quantization scheme (INT8 weight with INT8 activation) for \OURS.

\subsection{Layer-by-layer Knowledge Distillation with Affordable Cost}

Knowledge distillation (\kd) is one of the most powerful methods to alleviate the accuracy degradation after model compression. 
However, there are several limitations of \kd, especially for hidden-states \kd on large-scale language models: 
(1) \kd needs to hold a teacher and a student model together during the training, which dramatically increases the memory and compute cost;
(2) \kd usually requires full training of the student model. 
Therefore, several copies (gradient, first/second order momentum) of the weight parameters need to be stored in memory to update the model;
(3) \kd generally requires original training data, which sometimes are not accessible due to  privacy/confidential issues.

To address those limitations, we present our layer-by-layer distillation (\lwd) algorithm. 
Assume the target model for quantization has $N$ transformer blocks, $L_1$, ..., $L_N$, the accessible dataset has input $(\mX,~\mY)$, which can be the original training data or datasets from other resources. 
Our \lwd quantizes the network layer-by-layer and uses its original (i.e., unquantized) version as the teacher model. 
More specifically, assume layer $L_k$ is going to be quantized, and its quantized version is $\widehat L_k$. 
Then we use the output of the $L_{k-1}$ (i.e., by running inference on $X$ over the first $k-1$ layers) 
as the input of $L_k$ and $\widehat L_k$, measure the difference, and do the model update to $L_k$, i.e.,
\begin{equation}
\label{eq:lwd_loss}
\small
    \loss_{\lwd, k} = MSE\left(L_k \cdot L_{k-1} \cdot L_{k-2} \cdot ... \cdot L_{1} (\mX) - \widehat L_k \cdot L_{k-1} \cdot L_{k-2} \cdot ... \cdot L_{1} (\mX)\right),
\end{equation}
where $MSE$ is the mean square loss, and it can be also replaced by other losses (e.g., KL divergence) as well. 
As can be seen, 
(1) our \lwd does not need to hold a separate teacher as we use the same $L_1$ to $L_{k-1}$ for both teacher/student model. As such, the only extra model cost we have is $L_{k}$; 
(2) the memory overhead of optimizer states are significantly reduced as the only optimizing layer is $L_k$; 
(3) as we never optimize the end-to-end model, the training does not depend on the label anymore. 
Later, we will show that \lwd does not rely on the original training data in~\sref{sec:no_original_training_data}.

\subsection{Quantization-Optimized Transformer Kernels}
\label{sec:system_optimization}

\begin{figure}[t]
\centering
\includegraphics[width=0.99\linewidth]{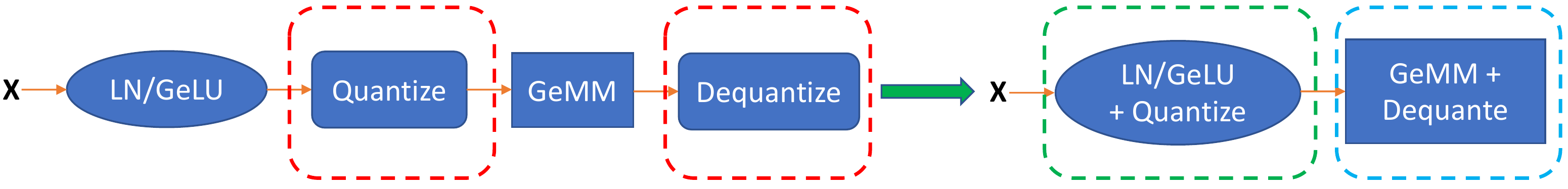}
\caption{
The illustration of normal (left) and our fused (right) INT8 GeMM.
}
\label{fig:operator_fusion}
\end{figure}

Both optimizing the inference latency and model size is crucial for serving large-scale transformer models in practice. 
During inference, the batch size is often relatively small, so the inference latency of the model primarily depends on the time of loading inference needed data from the main memory.
By quantizing the weights and activations to lower precision, we reduce the data volume needed to load those data, which allows more effective use of memory bandwidth and higher loading throughput.
However, simply converting weights/activations to INT8 does not guarantee improved latency because there are additional data movement overhead associated with quantization/dequantization operations as shown in~\fref{fig:operator_fusion} (red box). 
Such an overhead becomes expensive and in some cases surpasses the performance benefits of using low precision.
To reap the accuracy improvement from token-wise quantization while obtaining improved latency, we now present our optimizations that maximize the memory bandwidth utilization to speed up inference latency for \OURS.

\noindent
\textbf{CUTLASS INT8 GeMM} To support INT8 computation, we use CUTLASS~\cite{cutlass} INT8 GeMM implementation tuned for different batch sizes. 
Unlike standard GPU backend library, such as cuDNN, using CUTLASS allows us to more flexibly fuse quantization operation before and after GeMM to reduce kernel launching and data-movement overhead.

\noindent
\textbf{Fusing Token-wise Activation Quantization} Token-wise quantization/dequantization introduce many additional operations that lead to extra data movement cost. 
To eliminate these cost, we use \emph{kernel fusion}~\cite{kernel-fusion} to fuse quantization operation for activation with its previous element-wise and/or reduction operations such as bias-add, GeLU, and LayerNorm into a single operator, as illustrated by the green box in~\fref{fig:operator_fusion}. 
For the dequantization operation (e.g., dequantizing the integer output from the GeMM operator), we similarly fuse it with our custom GeMM schedule to avoid additional read/write accesses to the main memory as illustrated by the blue box in~\fref{fig:operator_fusion}.

By doing the above optimizations,
we are able to show significant latency reduction for \bert and \gpt-style models in~\sref{sec:results}.
Please see~\appref{sec:details_about_system_optimization} for more details about our system optimization. 

\begin{table}[t]
\caption{
Result of \bertbase on the development set of GLUE benchmark (except WNLI). 
\cite{shen2020q}$^+$ uses 128 groups for weight matrix which is hard to get GPU acceleration.
\cite{bondarenko2021understanding}$^*$ uses mixed INT8 and FP16 activation, and it directly reports the average metric of MNLI/MRPC/QQP/STS-B, which is basically the average of the two metrics we used for our runs.
}\centering
\label{tab:bert_base_main_result}
\begin{adjustbox}{width=0.99\linewidth}
\centering
\begin{tabular}{lcccccccccccccc }
\toprule
Precision (Method)   & CoLA  & MNLI-m & MNLI-mm & MRPC & QNLI & QQP & RTE & SST-2 & STS-B    & Ave. & Ave. Time (s)\\
\midrule
W16A16 (Baseline)   & 59.72 & 84.94 & 85.06 & 86.27/90.57 & 92.15 & 91.51/88.56 & 72.20 & 93.23 & 90.06/89.59 & 83.95 & N/A \\
\midrule
W8A8~\cite{shen2020q} (\qat)$^+$ & --- & 83.91 & 83.83 & --- & --- & --- & --- & 92.83 & --- & --- & --- \\
W8A8~\cite{zafrir2019q8bert} (\qat) & 58.48 & --- & --- & ---/89.56 & 90.62 & ---/87.96 & 68.78 & 92.24 & 89.04/--- & --- & ---\\
W8A8 (\qat)  & 61.21 & 84.80 & 84.64 & 83.82/88.85 & 91.29 & 91.29/88.28 & 71.12 & 92.89 & 88.39/88.18 & 83.37 & 2900 \\
W8A8 (\ptq)         & 56.06 & 79.99 & 81.06 & 75.49/79.67 & 87.35 & 89.92/86.82 & 48.38 & 91.40 & 86.58/86.44 & 77.41 & 6 \\
W8A8/16~\cite{bondarenko2021understanding} (\ptq)$^*$ & 58.63 & 82.67 & 82.67 & 88.74 & 90.41 & 89.40 & 68.95 & 92.66 & 88.00 & 82.46 & Unknown \\
W8A8 (\OURS)       & 59.59 & 84.83 & 85.13 & 86.03/90.39 & 91.98 & 91.45/88.46 & 71.12 & 93.12 & 90.09/89.62 & 83.75 & 0 \\
\midrule
W4/8A16 (\ptq)      & 0.00 & 16.74 & 16.95 & 31.62/0.00 & 50.74 & 63.18/0.00 & 47.29 & 70.64 & 16.48/15.91 & 33.11 & 6 \\
W4/8A16 (\OURS)    & 57.29 & 82.69 & 83.27 & 84.56/88.40 & 90.04 & 86.52/79.49 & 70.76 & 92.78 & 88.46/88.61 & 81.65 & 0 \\
W4/8A16 (\OURSlwd)     & 58.50 & 83.16 & 83.69 & 84.80/89.31 & 90.83 & 88.94/84.12 & 70.04 & 92.78 & 88.49/88.67 & 82.35 & 31 \\
\midrule
W4/8A8 (\OURS)     & 56.69 & 82.46 & 83.06 & 84.07/88.03 & 90.13 & 87.04/80.50 & 70.76 & 92.78 & 88.07/88.44 & 81.55 & 0\\
W4/8A8 (\OURSlwd)      & 58.80 & 83.09 & 83.65 & 85.78/89.90 & 90.76 & 89.16/84.85 & 71.84 & 93.00 & 88.16/88.55 & 82.71 & 31 \\
\bottomrule
\end{tabular}
\end{adjustbox}
\end{table}

\section{Results}
\label{sec:results}

\noindent
\textbf{Experimental Details}
To evaluate the proposed \OURS, we test it on both BERT and \gpt models. 
For BERT, we tested both \bertbase and \bertlarge on GLUE benchmark; 
and for \gpt-style models, we tested the \gpttf (i.e., \gpt-style model with 350M parameters) and \gptot (i.e., \gpt-style model with 1.3B parameters) on 20 zero-shot evaluation tasks, including 19 accuracy-based tasks and 1 language modeling generation task. 
To illustrate the scalability of the proposed \OURS, we also directly apply it to two of the largest open-sourced \gpt-style models, i.e., \gptj~\cite{gpt-j} and \gptneox~\cite{gpt-neox}. 
We use a fixed set of hyperparameters for all the \lwd-related experiments even though tuning them may benefit our results. 
Please see~\appref{sec:main_result_training_details} for more training details and see~\appref{sec:accuracy_reported_for_bert_on_glue} for the reported metrics for \bert.
To provide a comprehensive study, we also include a tuning result in~\appref{sec:tuning_results_on_bert} on BERT and an ablation study for different proposed components in~\sref{sec:ablation_study_of_different_components}.

\noindent
\textbf{Notation Explanation}
We use WxAy to represent using x-bit for weight quantization and y-bit for activation quantization.
Unless specific explanation, for W4/8, we quantize the \mhsa's weight to INT8 and \ffc's weight to INT4; 
for A8/16, we use FP16 activation for self-attention calculation (i.e., the GeMM related to $\mW_{q/k/v}$) and use INT8 for the rest calculation. 
We use \OURS to represent the method with only fine-grained quantization schemes and use \OURSlwd to represent the method with both fine-grained quantization schemes and \lwd. 

\subsection{Main Results of BERT}

\noindent
\textbf{\bertbase}
We report the results of \bertbase in~\tref{tab:bert_base_main_result}. 
For W8A8, the average accuracy of \ptq degrades more than 10 points. 
However, \OURS can achieve 83.75 scores, which is only 0.2 lower than baseline. 
Particularly, as \OURS has no activation range calibration phase, the cost of \OURS is $0$ which is even cheaper than standard \ptq. 
As compared to~\cite{bondarenko2021understanding}, our method achieves a better average score (1.29 higher). 
Meanwhile, as compared to INT8 activation used in \OURS, \cite{bondarenko2021understanding} uses mixed INT8 and FP16 activation. 

We also compare our method with our internal trained \qat and other \qat works~\cite{shen2020q,zafrir2019q8bert}. 
As can be seen, with comparable accuracy results as those \qat methods, \OURS can save the retraining cost from 2900s to 0s for INT8 quantization.

For the more aggressive weight quantization with minimal (or no) training quantization, i.e., W4/8A16, \ptq fully loses all accuracy (pure random prediction). 
However, \OURS can still achieve an 81.65 average score. 
On top of \OURS, if we add our \lwd, the accuracy can be further boosted to 82.35 with a cost of 31s per task using only a single GPU, which is 93.5x cheaper than INT8 \qat quantization.
We also test \OURS and \OURSlwd under the W4/8A8 quantization scheme and both of them achieve similar accuracy performance as W4/8A16. 
If hyper-parameter tuning is applied to \lwd, \OURSlwd can achieve an 83.22 average score under W4/8A8, which is similar to \qat's W8A8 result. 
Please see~\appref{sec:tuning_results_on_bert} for more details.

\noindent
\textbf{\bertlarge}
We test our methods on \bertlarge as well and the results are shown in~\tref{tab:bert_large_main_result}. 
Similar to \bertbase, \OURS achieves much better accuracy than \ptq methods. 
As compared to \qat methods, \OURS has comparable results on larger datasets (like MNLI/QQP) and has better performance on small tasks (e.e., CoLA/MRPC/RTE). 
We actually tune \qat for multiple learning rates but cannot get even better performance for those small tasks (see~\appref{sec:qat_for_bearlarge} for more details). 

For more aggressive quantization schemes, like W4/8A16 and W4/8A8, \OURS and \OURSlwd still achieve good accuracy except for RTE but the model size is about 3x smaller than FP16 counterpart. 
This is aligned with the INT8 \qat results, which lose significantly more accuracy on RTE. 
Thanks to the lightweight cost of \lwd, it only takes about 550s to finish each task even on \bertlarge, which is 13x cheaper than \qat.

\begin{table}[t]
\caption{
Result of \bertlarge on the development set of GLUE benchmark (except WNLI). 
$^+$We extensively tuned the learning rate for \qat (see~\appref{sec:qat_for_bearlarge} for more details).
}
\centering
\label{tab:bert_large_main_result}
\begin{adjustbox}{width=0.99\linewidth}
\centering
\begin{tabular}{lcccccccccccccc }
\toprule
Precision (Method)   & CoLA  & MNLI-m & MNLI-mm & MRPC & QNLI & QQP & RTE & SST-2 & STS-B    & Ave. & Ave. Time (s)\\
\midrule
W16A16 (Baseline)   & 63.35 & 86.65 & 85.91 & 87.99/91.62 & 92.24 & 91.08/88.08 & 74.01 & 93.46 & 90.34/90.11 & 85.03 & N/A \\
\midrule
W8A8~\cite{zafrir2019q8bert} (\qat) & --- & --- & --- & ---/90.9 & 91.74 & & & & 90.12/--- & --- & ---\\
W8A8 (\qat)$^+$ & 59.85 & 86.65 & 86.35 & 85.29/89.43 & 92.55 & 91.60/88.60	& 61.37 & 93.23 & 87.55/87.65 & 82.78 & 7181 \\
W8A8 (\ptq)         & 60.57 & 75.69 & 76.94 & 81.13/84.93 & 88.49 & 84.04/74.35 & 46.93 & 91.74 & 62.75/55.77 & 73.54 & 31 \\
W8A8 (\OURS)       & 63.38 & 86.52 & 85.64 & 87.75/91.50 & 92.31 & 91.09/88.05 & 72.56 & 93.35 & 90.45/90.19 & 84.81 & 0 \\
\midrule
W4/8A16 (\ptq)      & 0.00 & 16.85 & 33.24 & 68.38/80.89 & 51.25 & 63.18/0.00 & 52.71 & 52.41 & -5.74/-8.51 & 35.73 & 31 \\
W4/8A16 (\OURS)    & 62.99 & 84.77 & 84.42 & 87.50/91.16 & 91.63 & 90.03/86.41 & 48.01 & 92.16 & 89.49/89.28 & 81.23 & 0 \\
W4/8A16 (\OURSlwd)     & 63.72 & 84.90 & 84.81 & 87.99/91.39 & 91.45 & 90.34/86.92 & 51.62 & 92.43 & 89.46/89.29 & 81.85 & 550 \\
\midrule
W4/8A8 (\OURS)     & 62.34 & 84.62 & 84.25 & 87.75/91.38 & 91.87 & 89.86/86.09 & 47.65 & 91.97 & 89.39/89.17 & 81.06 & 0 \\
W4/8A8 (\OURSlwd)      & 63.51 & 84.70 & 84.71 & 88.73/91.99 & 91.73 & 90.25/86.74 & 49.82 & 92.09 & 89.34/89.08 & 81.62 & 550 \\
\bottomrule
\end{tabular}
\end{adjustbox}
\end{table}

\begin{table}[t]
\caption{
Post training quantization result of \gpttf on 20 zero-shot evaluation datasets. 
Please see~\tref{tab:gpt3_350_full_table} for the results of all 20 tasks.
}\centering
\label{tab:gpt3_350_main_result}
\begin{adjustbox}{width=0.99\linewidth}
\centering
\begin{tabular}{lcccccccccccccc }
\toprule
Precision (Method)     & Lambada ($\uparrow$)    & PIQA ($\uparrow$) & OpenBookQA ($\uparrow$) & RTE ($\uparrow$) & ReCoRd ($\uparrow$) & Ave. 19 Tasks ($\uparrow$) & \wikitext ($\downarrow$) & Time Cost\\
\midrule
W16A16     & 49.3 & 66.3 & 29.4 & 53.8 & 75.1 & 38.9 & 21.5  & N/A \\
\midrule
W8A8 (\ptq)       & 42.6 & 64.1 & 28.0 & 53.1 & 67.5 & 37.8 & 26.2 & 7 mins \\
W8A8 (\OURS)    & 51.0 & 66.5 & 29.2 & 53.4 & 74.9 & 38.7 & 21.7 & 0 \\
\midrule 
W4/8A16 (\ptq)       & 0.00 & 51.4 & 30.2 & 52.7 & 16.1 & 28.9 & 1.76e5 & 7 mins \\
W4/8A16 (\OURS)    & 10.1 & 58.5 & 27.2 & 52.0 & 56.5 & 33.5 & 88.6   & 0 \\
W4/8A16 (\OURSlwd)     & 39.8 & 63.8 & 29.4 & 53.1 & 70.1 & 37.0 & 30.6   & 1.1 hours \\ 
\midrule 
W4/8A8 (\OURS)     & 10.5 & 57.7 & 28.0 & 52.7 & 55.3 & 33.4 & 92.1  & 0 \\
W4/8A8 (\OURSlwd)      & 37.4 & 61.8 & 28.2 & 53.1 & 68.5 & 36.6 & 31.1 & 1.1 hours \\  
\bottomrule
\end{tabular}
\end{adjustbox}
\end{table}

\subsection{Main Results of \gpt-style Models}

\noindent
\textbf{\gpttf}
We first test \OURS and \OURSlwd on \gpttf and report the result in~\tref{tab:gpt3_350_main_result}. 
The first interesting finding of zero-shot evaluation on \gpt-stype models is that the accuracy performance of accuracy-based tasks is more tolerant to quantization than generation tasks. 
For instance, W8A8 \ptq has a 1.1\% average accuracy drop on 19 accuracy-based tasks as compared to 4.7 points loss on \wikitext. 
Comparing \OURS with \ptq using W8A8, we can reduce the accuracy gap from 1.1\% to 0.2\% and the perplexity (PPL) gap from 4.7 to 0.2 with no activation range calibration cost. 

For W4/8A16 quantization scheme, \ptq can hardly predict reasonable answers for the majority of tasks and its generation performance on \wikitext is fully crashed. 
As a comparison, \OURS still achieves non-trivial performance on some tasks but its generation performance significantly degrades on \wikitext. 
\lwd brings a significant performance boost for this W4/8A16 setting. 
Note that \OURSlwd increases the accuracy from 33.5 to 37.0 and decreases the PPL from 88.6 to 30.6 compared to \OURS, 
and the entire cost of this is just 3.1 hours on a single A100 GPU. 
Note that this is about 0.027\% GPU hours of the full pretraining cost (128 A100 GPUs for 32 hours).
Similar to W4/8A16, \OURSlwd achieves much better performance than \OURS on W4/8A8 by using the lightweight \lwd.

\begin{table}[t]
\caption{
Post training quantization result of \gptot on 20 zero-shot evaluation datasets.
Please see~\tref{tab:gpt3_13_full_table} for the results of all 20 tasks.
}\centering
\label{tab:gpt3_13_main_result}
\begin{adjustbox}{width=0.99\linewidth}
\centering
\begin{tabular}{lcccccccccccccc }
\toprule
Precision (Method)     & Lambada ($\uparrow$)    & PIQA ($\uparrow$) & OpenBookQA ($\uparrow$) & RTE ($\uparrow$) & ReCoRd ($\uparrow$) & Ave. 19 Tasks ($\uparrow$) & \wikitext ($\downarrow$) & Time Cost\\
\midrule
W16A16     & 61.3 & 71.4 & 33.6 & 53.1 & 82.6 & 42.4 & 15.3  & N/A \\
\midrule
W8A8 (\ptq)       & 54.8 & 67.7 & 16.6 & 54.5 & 75.7 & 40.5 & 18.9 & 13 mins \\
W8A8 (\OURS)     & 62.6 & 70.7 & 33.4 & 52.7 & 80.9 & 42.3 & 15.7   & 0 \\
\midrule 
W4/8A16 (\ptq)      & 0.00 & 50.4 & 27.0 & 50.9 & 15.8 & 29.0  & 1.35e5 &  13 mins\\
W4/8A16 (\OURS)    & 43.9 & 66.5 & 30.0 & 52.7 & 77.3 & 39.38 & 21.9   & 0 \\
W4/8A16 (\OURSlwd)     & 59.4 & 69.5 & 31.6 & 52.7 & 79.7 & 41.5  & 17.6   &  3 hours \\ 
\midrule 
W4/8A8 (\OURS)     & 46.8 & 66.4 & 28.8 & 52.7 & 76.2 & 39.24 & 24.1 & 0 \\
W4/8A8 (\OURSlwd)      & 48.7 & 68.1 & 29.0 & 52.0 & 77.4 & 39.90 & 18.2 & 3 hours \\  
\bottomrule
\end{tabular}
\end{adjustbox}
\end{table}

\begin{table}[t]
\caption{
The speedup of our W8A8 as compared to W16A16. 
We measure the end-to-end average latency for the entire \bert model, and the time reported is in milliseconds.
}\centering
\label{tab:bert_speedup}
\begin{adjustbox}{width=0.99\linewidth}
\centering
\begin{tabular}{lcccccccccccccccccccccccccccccccccccc}
\toprule
Seq Len & \multirow{2}{*}{Precision} & \multicolumn{8}{c}{128} & & \multicolumn{8}{c}{256}    \\
\cline{3-10} \cline{12-19}
BS      &  & 1 & 2 & 4 & 8 & 16 & 16 & 64 & 128 & & 1 & 2 & 4 & 8 & 16 & 16 & 64 & 128 \\
\midrule
\multirow{3}{*}{\bertbase} & W16A16  & 2.45  & 3.22  & 3.85  & 5.51  & 9.96  & 17.93 & 34.25 & 67.08 & & 3.13  & 4.05  & 5.70  & 10.55 & 19.27 & 36.69 & 71.75 & 140.0 \\
                           & W8A8    & 1.08  & 1.16  & 1.42  & 1.76  & 2.58  & 3.90  & 6.74  & 12.92 & & 1.22  & 1.44  & 2.08  & 2.88  & 4.10  & 7.80  & 14.66 & 28.13 \\
                           & Speedup & 2.27  & 2.78  & 2.71  & 3.13  & 3.86  & 4.60  & 5.08  & 5.19  & & 2.57  & 2.81  & 2.74  & 3.66  & 4.70  & 4.70  & 4.89  & 4.98 \\
\midrule
\multirow{3}{*}{\bertlarge} & W16A16  & 5.45  & 6.38  & 8.73  & 13.88 & 26.34 & 48.59 & 92.49 & 183.4 & & 6.39  & 8.94  & 14.66 & 27.99 & 51.94 & 98.78 & 195.9 & 384.5 \\
                            & W8A8    & 2.08  & 2.58  & 2.84  & 3.79  & 6.21  & 10.28 & 18.86 & 36.62 & & 2.55  & 3.36  & 4.16  & 6.88  & 11.61 & 21.20 & 41.24 & 79.90 \\
                            & Speedup & 2.62  & 2.47  & 3.07  & 3.66  & 4.24  & 4.73  & 4.90  & 5.01  & & 2.51  & 2.66  & 3.52  & 4.07  & 4.47  & 4.66  & 4.75  & 4.81  \\
\bottomrule
\end{tabular}
\end{adjustbox}
\end{table}

\noindent
\textbf{\gptot}
The results of \gptot are shown in~\tref{tab:gpt3_13_main_result}. 
Similar to \gpttf, for W8A8, \OURS has much better performance than \ptq with less no activation calibration cost, particularly for the generation task \wikitext (3.2 points lower). 
Also, for W4/8 quantization, \lwd can bring non-trivial performance gain for \OURS. 
The cost of \lwd is about 0.02\% of the full pre-training cost (128 A100 GPUs for 120 hours)

\subsection{Latency Reduction of BERT and \gpt-style Models}
We compare the inference speed of BERT between FP16 and our INT8 versions in~\tref{tab:bert_speedup} on a single 40G-A100 GPU.
Using our efficient quantization kernel implementation and operator fusion, the INT8 model can achieve 2.27--5.19x speedup on \bertbase and 2.47--5.01x on \bertlarge.

We also include the latency comparison of \gpt-style models between FP16 and our INT8 version. 
Particularly, we use the model to generate the first 50 tokens based on a given text and measure the average latency. 
Our INT8 model leads to 4.16x/4.06x speedup for \gpttf/\gptot as compared to the FP16 counterpart.

\subsection{A Showcase of \gptj and \gptneox}
To demonstrate the scalability of \OURS, we applied it to two of the largest open-sourced models, i.e., \gptj and \gptneox, which have 6B and 20B parameters separately. 

We report the results of \gptj in~\tref{tab:gptj_huggingface} on three generation datasets, i.e., PTB~\cite{marcinkiewicz1994building}, \wikitext, and \wikitextone~\cite{merity2016pointer}. 
As can be seen, as compared to FP16 precision, \OURS achieves similar PPL on all three different tasks. 
To compare the latency, we again use the average latency number to generate the first 50 tokens.
Our W8A8 can get up to 3.67x speedup compared to the FP16 version.

\begin{table}[t]
\begin{minipage}[t]{.495\linewidth}
    \caption{
Post training quantization result of \gptj on three zero-shot generation tasks
}\centering
\setlength\tabcolsep{1.5pt}
\label{tab:gptj_huggingface}
\begin{adjustbox}{width=0.99\linewidth}
\centering
\begin{tabular}{lcccccccccccccc }
\toprule
Precision & PTB & \wikitext & \wikitextone & Latency\\
\midrule
W16A16 & 20.47 & 10.35 & 10.35 & 29.13ms (1x)\\
W8A8   & 20.97 & 10.51 & 10.52 & 7.94ms (3.67x)\\
\bottomrule  
\end{tabular}
\end{adjustbox}
\end{minipage}\hfill
\begin{minipage}[t]{.495\linewidth}
\caption{
Post training quantization result of \gptneox on 19 zero-shot evaluation datasets.
Please see~\tref{tab:gptneox_full_table} for the results of all 19 tasks. 
}\centering
\setlength\tabcolsep{1.5pt}
\label{tab:gptneox_main_text}
\begin{adjustbox}{width=0.99\linewidth}
\centering
\begin{tabular}{lcccccccccccccc }
\toprule
Precision  & Lambada   & PIQA & Ave. 19 Tasks & Latency \\
\midrule
W16A16  & 71.7 & 77.7 & 50.5 & 2$\times$65ms (1x)\\
W8A8/16 & 71.9 & 78.3 & 50.4 & 1$\times$25ms (5.2x)\\
\bottomrule
\end{tabular}
\end{adjustbox}
\end{minipage}\hfill
\end{table}

To quantize \gptneox to W8A8 for all GeMMs, the accuracy significantly decreases. 
We retrieve the quantization of each weight matrix and of each activation, and finally find out that the activation quantization for the attention calculation (i.e., the input of self-attention) causes the accuracy loss. 
We conjecture that this is because of the sensitivity of the self-attention module for extra-large models (20B) but cannot verify this for other models due to the lack of open-sourced extra-large models and the full evaluation pipeline. 
As such, we leave the input activation for self-attention in FP16 and quantize the rest to INT8.
The results are shown in~\tref{tab:gptneox_main_text}. 
Our W8A8/16 achieves similar accuracy performance but can reduce both the GPU resource requirement (from 2 A100 GPUs to 1) and the latency from 65ms to 25ms, 
which together lead to 5.2x better throughput/efficiency.

\subsection{Ablation Study of Different Components}
\label{sec:ablation_study_of_different_components}

To investigate the performance gain of each component we introduced in~\sref{sec:methodology}, i.e., group-wise weight quantization, token-wise activation quantization, and lightweight layer-by-layer knowledge distillation, we here do an ablation study on \bertlarge with W4/8A8. 

We present the results in~\tref{tab:bert_large_ablation_study_component}.
As can be seen, group-wise weight quantization boosts the accuracy (random-guess prediction) from \ptq to a non-trivial result (66.52). 
Further adding token-wise quantization improves 14.54 points accuracy performance. 
On top of those (i.e., \OURS), \lwd further brings a 0.56 point gain. 

\begin{table}[t]
\caption{
Ablation study of different components for \bertlarge on the development set of GLUE. 
The quantization scheme used here is W4/8A8.
Here, GP is the abbreviation of group-wise weight quantization, TQ is the abbreviation of token-wise activation quantization.
}\centering
\label{tab:bert_large_ablation_study_component}
\begin{adjustbox}{width=0.99\linewidth}
\centering
\begin{tabular}{lccccccccccccccccc }
\toprule
GQ     & TQ     & \lwd   & CoLA  & MNLI-m & MNLI-mm & MRPC & QNLI & QQP & RTE & SST-2 & STS-B    & Ave.\\
\xmark & \xmark & \xmark & -0.79 & 33.07 & 32.94 & 68.38/80.54 & 49.42 & 63.18/0.00 & 52.71 & 52.29 & -4.27/-1.90 & 35.85 \\
\cmark & \xmark & \xmark & 59.81 & 66.63 & 68.79 & 68.63/71.17 & 83.87 & 78.24/61.30 & 46.93 & 89.45 & 54.58/32.52 & 66.52 \\
\cmark & \cmark & \xmark & 62.34 & 84.62 & 84.25 & 87.75/91.38 & 91.87 & 89.86/86.09 & 47.65 & 91.97 & 89.39/89.17 & 81.06 \\
\cmark & \cmark & \cmark & 63.51 & 84.70 & 84.71 & 88.73/91.99 & 91.73 & 90.25/86.74 & 49.82 & 92.09 & 89.34/89.08 & 81.62 \\
\bottomrule
\end{tabular}
\end{adjustbox}
\end{table}

\begin{table}[!t]
\caption{
Post training quantization result of \gpttf on 20 zero-shot evaluation datesets
The quantization scheme here is W4/8A8.
Please see~\tref{tab:gpt3_350_data_ablation_study_full_table} for the results of all 20 tasks.
}\centering
\label{tab:gpt3_350_data_abalation_study}
\begin{adjustbox}{width=0.99\linewidth}
\centering
\begin{tabular}{lcccccccccccccc }
\toprule
Method & Data Resource    & Lambada ($\uparrow$)    & PIQA ($\uparrow$) & OpenBookQA ($\uparrow$) & RTE ($\uparrow$) & ReCoRd ($\uparrow$) & Ave. 19 Tasks ($\uparrow$) & \wikitext ($\downarrow$) \\
\midrule
\OURS   & ---             & 10.5 & 57.7 & 28.0 & 52.7 & 55.3 & 33.4 & 92.1  \\
\OURSlwd   & Random data     & 26.1 & 59.3 & 29.2 & 50.5 & 64.9 & 34.5 & 40.6  \\
\OURSlwd   & Wikipedia       & 33.9 & 62.4 & 28.0 & 52.7 & 69.5 & 36.2 & 30.4  \\
\OURSlwd   & Original data   & 37.4 & 61.8 & 28.2 & 53.1 & 68.5 & 36.6 & 31.1  \\  
\bottomrule
\end{tabular}
\end{adjustbox}
\end{table}

\subsection{No Access to The Original Training Data}
\label{sec:no_original_training_data}

As mentioned in previous sections, the original training data are oftentimes hard to access due to the privacy and/or confidential issues. 
Therefore, we here study the performance of our \lwd when there is no direct access to the original training data. 
As the distillation objective of our \lwd does not depend on the label, the training data used for \lwd can be very flexible. 

We compare the performance of \gpttf on W4/8A8 quantization scheme using three different training data resources, i.e., random data (using random integer number to generate token ids), Wikipedia (using Huggingface to get the data\footnote{\url{https://huggingface.co/datasets/wikipedia}}), and original PILE dataset. 

The results are shown in~\tref{tab:gpt3_350_data_abalation_study}. 
Compared to \OURS, \lwd using random data can boost the accuracy by 1.1\% and reduce the PPL from 92.1 to 40.6. 
The reason why random data can still significantly improve the performance is that \lwd does not optimize the end-to-end pipeline and it only layer-by-layer learns the internal dependency from the teacher model. 
Therefore, random data can also provide meaningful information.
Using Wikipedia data from Huggingface can further improve the accuracy to 36.2 and reduce the PPL to 30.4, which is comparable to the results using the original data. 
This indicates that a clean text dataset can be used for \lwd when we do not have access to the original full dataset.

\section{Conclusions}
\label{sec:conclusions}

With the rapid growth of large model sizes, we have reach a point to consider how to serve those models in practice. 
Although several works demonstrate that post-training quantization can be applied to \bert models, to the best of our knowledge, there have been no existing works on (1) billion-scale \gpt-style models, (2) ultra-low precision post-training quantization, and (3) end-to-end solution of how to efficiently serve the quantized model online. 
In this work, we offer fine-grained compression schemes for both weight and activations to enable INT8 quantization for up to 20B-scale models (\gptneox). 
We also offer a novel affordable layer-by-layer knowledge distillation for ultra-low precision quantization, which leads to 3x model size reduction compared to FP16 model while achieving minimal accuracy degradation. 
Furthermore, we provide a system backend support and show up to 5.19x speedup on \bert models and 5.2x better efficiency on \gptneox.

\section*{Acknowledgments}
This work is done within the DeepSpeed team in Microsoft. We appreciate the help from the DeepSpeed team. Particularly, we thank Jeff Rasley and Elton Zheng for solving the engineering issue. We thank the engineering supports from the Turing team in Microsoft.
\clearpage

{
\bibliographystyle{plain}
\bibliography{ref.bib}
}

\clearpage
\onecolumn
\appendix

\counterwithin{figure}{section}
\counterwithin{table}{section}

\section{Background}
\label{sec:background}
\subsection{Transformer Architecture}

The transformer architecture usually has three components: an embedding layer, a stack of encoder/decoder layers, and a final classifier. 
In this paper, we focus on quantizing the encoder/decoder layers, i.e., the transformer block, because it is often the most memory and compute intensive components in the entire architecture. With a transformer block, there are two sub-layers, the multi-head self-attention (\mhsa) and the feed-forward connection (\ffc). 
We give a short review later and please refer to~\cite{vaswani2017attention} for more details. At high level, transformer models can be broadly categorized to three branches: encoder-only models (\bert)~\cite{tenney2019bert}, decoder-only models (\gpt-style)~\cite{radford2019gpt}, and encoder-decoder models (T5)~\cite{colin2019t5}. 
In this paper, we focus on encoder-only and decoder-only models but our approach can be applied to encoder-decoder models as well.

\begin{figure}[h]
\centering
\includegraphics[width=0.99\linewidth]{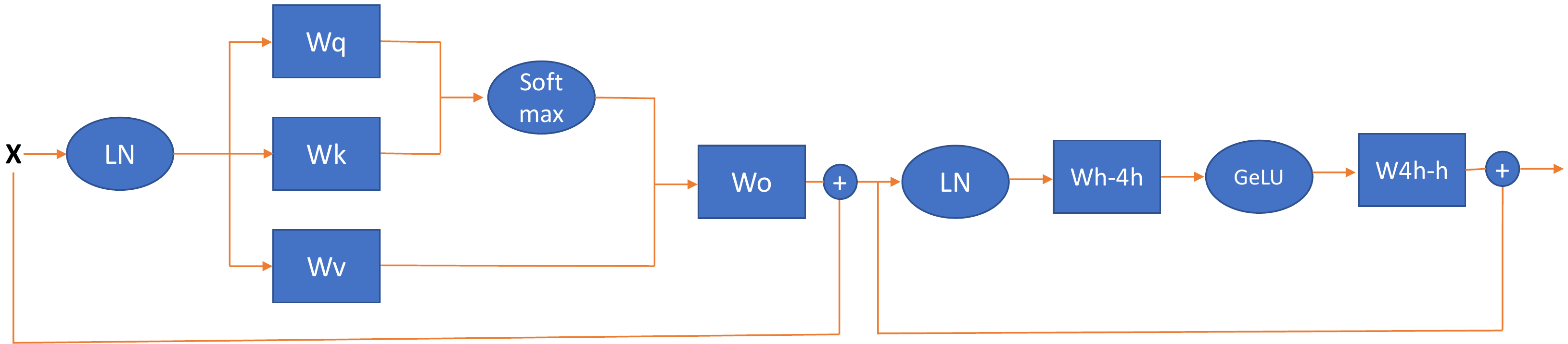}
\caption{
The illustration of a Transformer-block. 
}
\label{fig:transformer_block_illustration}
\end{figure}

\paragraph{Transformer Block}
Assume the input of an encoder layer is $\mX$, 
the query, key, value, attention output, \ffc dense, and \ffc output matrices are $\mW_q$, $\mW_k$, $\mW_v$, $\mW_o$, $\mW_{h-4h}$, and $\mW_{4h-h}$, respectively. 
Then the forward propagation of a transformer-block is illustrated in~\fref{fig:transformer_block_illustration}, where LN is the layer normalization, Softmax is the softmax operator, and GeLU is the activation function.

\subsection{Quantization Background}
\label{sec:quantization_background}
Quantization maps high-precision numbers, e.g., FP16/FP32, to its low-precision counterpart, e.g., INT4/INT8, to reduce the model footprint and improve the compute performance. 
In this work, we use uniform symmetric scalar quantizers. 
That is to say, if we have a vector/matrix, $\rvx$, the quantization is applied as
\begin{equation}
    \rvx_{quantize} = round\left(clamp(\frac{\rvx}{S}, -2^{bit-1}, 2^{bit-1}-1)\right),
\end{equation}
where $bit$ is the number of bit we use to represent the quantized value, and $S$ is the scaling factor. 
For weight matrix quantization, $S$ is generally computed as
$S = max\left( abs(\rvx) \right)$,
since the weight matrix is static during inference. 
On the other hand, activations' range is dynamic during inference so that an accurate $S$ requires dynamic calculation during inference. 
However, to achieve best latency reduction, coarse-grained static quantization is usually applied in practice, where $S$ is calibrated using training data (e.g., momentum based averaging) and fixed during inference~\cite{gholami2021survey}. 
Although static quantization achieves better latency reduction, it also limits the quantization representation for activations, which is discussed in~\sref{sec:ptq_challenge}.

\section{Experimental Details}
\label{sec:all_experimental_details}
\subsection{Details of \ptq on BERT and GPT}
\label{sec:ptq_calibrate}
For BERT, we use a batch size of 32 and sequence length 128 to calibrate the range of activations. 
In order to capture the dynamic range, we use 0.95 momentum with 100 iterations, i.e., 
\begin{align*}
        x_{max} = 0.95x_{max} + 0.05max(x_{current-iteration}), \\
        x_{min} = 0.95x_{min} + 0.05min(x_{current-iteration}).
\end{align*}
For \gpt-style models, we use the same momentum method but change the batch size to 8 with sequence length 2048.

\subsection{Details of Main Result}
\label{sec:main_result_training_details}
\paragraph{BERT}
BERT models are trained using the code-base from Huggingface~\cite{wolf2019huggingface}. 
We show our \OURS method on \bertbase and \bertlarge.
We use the same lower-case tokenizer in \bertlarge instead of the cased tokenizer in the original paper~\cite{devlin2018bert}.
When fine-tuning on GLUE~\cite{wang2018glue} tasks ((i.e.,
MRPC~\cite{dolan2005automatically},
STS-B~\cite{cer2017semeval},
SST-2~\cite{socher2013recursive}, QNLI~\cite{rajpurkar2016squad}, QQP~\cite{iyer2017first}, MNLI~\cite{williams2018broad}, CoLA~\cite{warstadt2018neural}, RTE~\cite{dagan2013recognizing}).\footnote{We exclude WNLI~\cite{levesque2012winograd} since its results are not stable~\cite{dodge2020fine}.}), we follow the instruction from Huggingface Transformer Library~\cite{wolf2019huggingface}. 

For \OURS and \OURSlwd, we use 48 groups for group-wise weight quantization on \bertbase and 64 groups for group-wise weight quantization on \bertlarge, for all the weight matrices. 

For \lwd, we use 100 iterations with batch size 32 and sequence length 128 for \bertbase, and we use 400 iterations for \bertlarge. 
We fix the learning rate as 5e-6 for both models on all tasks. 
However, tuning them may favor \OURS. 

All the models are trained using a single 40G-A100 GPU (Azure ND A100 instances).

\paragraph{\gpt-style Models}
All \gpt-style models used in the paper are trained using DeepSpeed~\cite{rasley2020deepspeed} and Megatron-DeepSpeed Library~\footnote{\url{https://github.com/microsoft/Megatron-DeepSpeed}}. 
The pretraining data are from PILE dataset~\cite{gao2020pile}, and the training pipeline and hyperparameters are based on the Megatron-DeepSpeed repository. 
We use 128 A100 GPUs (Azure ND A100 instances) to do the pretraining.
It takes about 32 hours to finish the training of \gpttf and 120 hours of \gptot.
We evaluate our results on 20 zero-shot evaluation tasks, including 19 accuracy evaluation tasks (i.e., HellaSwag~\cite{zellers2019hellaswag}, LAMBADA~\cite{paperno2016lambada}, TriviaQA~\cite{joshi2017triviaqa}, WebQS~\cite{berant-etal-2013-semantic}, Winogrande~\cite{sakaguchi2020winogrande}, PIQA~\cite{tata2003piqa}, ARC (Challenge/Easy)~\cite{boratko2018systematic}, ANLI (R1/R2/R3)~\cite{williams2020anlizing}, OpenBookQA~\cite{mihaylov2018can}, RACE-h~\cite{lai2017race}, BoolQ~\cite{clark2019boolq}, Copa~\cite{afshar2018copa}, RTE~\cite{dagan2013recognizing}, WSC~\cite{levesque2012winograd}, MultiRC~\cite{yadav2019quick}, ReCoRD~\cite{zhang2018record}) and 1 language modeling generation task (i.e., \wikitext~\cite{merity2016pointer}). 

For \OURS and \OURSlwd, we use 64/128 groups for group-wise weight quantization on \gpttf/\gptot for all the weight matrices. 

For \lwd, we use 1600 iterations with batch size 8 and sequence length 2048 for both \gpttf and \gptot. 
We fix the learning rate as 5e-6 for both models. 
However, tuning them may favor \OURS. 

All the quantized models are trained using a single 40G-A100 GPU (Azure ND A100 instances).

\subsection{Accuracy reported for BERT on GLUE}
\label{sec:accuracy_reported_for_bert_on_glue}
We report the performance metric for BERT on GLUE based on~\tref{tab:bert_reported_numbers}.
For the average score, if the task only has one metric, we use it for the final result; if the task has two metrics, we compute the average of the two metrics first and use it for the final average score. 
For instance, the score of MRPC used to compute the final average is the mean of its accuracy and F1 score. 

\begin{table}[t]
\caption{
Metric used for \bertbase on the development set of GLUE benchmark (except WNLI). 
}\centering
\label{tab:bert_reported_numbers}
\begin{adjustbox}{width=0.99\linewidth}
\centering
\begin{tabular}{cccccccccccccc }
\toprule
& CoLA  & MNLI-m & MNLI-mm & MRPC & QNLI & QQP & RTE & SST-2 & STS-B    \\
\midrule
& Matthews Correction & Accuracy & Accuracy / F1 & Accuracy & Accuracy & Accuracy / F1 & Accuracy & Accuracy & Pearson / Spearmanr\\
\bottomrule
\end{tabular}
\end{adjustbox}
\end{table}

\section{\ptq challenge of \bertbase}
\label{sec:ptq_challenge_of_bert}
From~\tref{tab:bert_investigation}, we observe similar results as~\cite{bondarenko2021understanding}, where the accuracy degradation of INT8 quantization is mainly from activation quantization. 
Specifically, there is a negligible accuracy drop from INT8 weight quantization (i.e., the row of W8A16). 
However, with sole INT8 activation (i.e., the row of W16A8), the accuracy decreases from 84.06 to 79.61. 
Besides, we also push the weight quantization to a mixed-precision setting with INT4 for weights in \ffc and INT8 for weights in \mhsa (i.e., the row of W4/8A16). 
This ultra-low precision quantization leads the model to be purely random without meaning prediction.

\begin{table}[t]
\caption{
Post training quantization results of \bertbase on development sets of the GLUE benchmark (except WNLI). 
Here WxAy means x-bit for weight quantization and y-bit for activation quantization.
Particularly, for W4/8, we quantize the \mhsa's weight to INT8 and \ffc's weight to INT4. 
Please see~\appref{sec:accuracy_reported_for_bert_on_glue} for the reported metrics.
}\centering
\label{tab:bert_investigation}
\begin{adjustbox}{width=0.9\linewidth}
\centering
\begin{tabular}{lcccccccccccccc }
\toprule
 Precision     & CoLA  & MNLI-m & MNLI-mm & MRPC & QNLI & QQP & RTE & SST-2 & STS-B  & Ave. \\
\midrule
W16A16     & 59.72 & 84.94 & 85.06 & 86.27/90.57 & 92.15 & 91.51/88.56 & 72.20 & 93.23 & 90.06/89.59 & 83.95 \\
W8A16      & 60.77 & 84.65 & 84.92 & 85.29/89.86 & 91.84 & 91.52/88.56 & 71.84 & 93.46 & 89.89/89.50 & 83.87 \\
W16A8      & 56.85 & 80.55 & 81.48 & 84.07/89.33 & 91.34 & 91.30/88.07 & 68.59 & 93.46 & 88.74/88.74 & 81.93 \\
W8A8       & 58.74 & 79.99 & 81.06 & 84.31/89.51 & 91.18 & 91.24/88.03 & 70.76 & 92.66 & 88.33/88.73 & 82.16 \\
W4/8A16    & 0.00 & 16.74 & 16.95 & 31.62/0.00 & 50.74 & 63.18/0.00 & 47.29 & 70.64 & 16.48/15.91 & 33.11 \\
\bottomrule
\end{tabular}
\end{adjustbox}
\end{table}

\begin{figure}[h]
\centering
\includegraphics[width=0.49\linewidth]{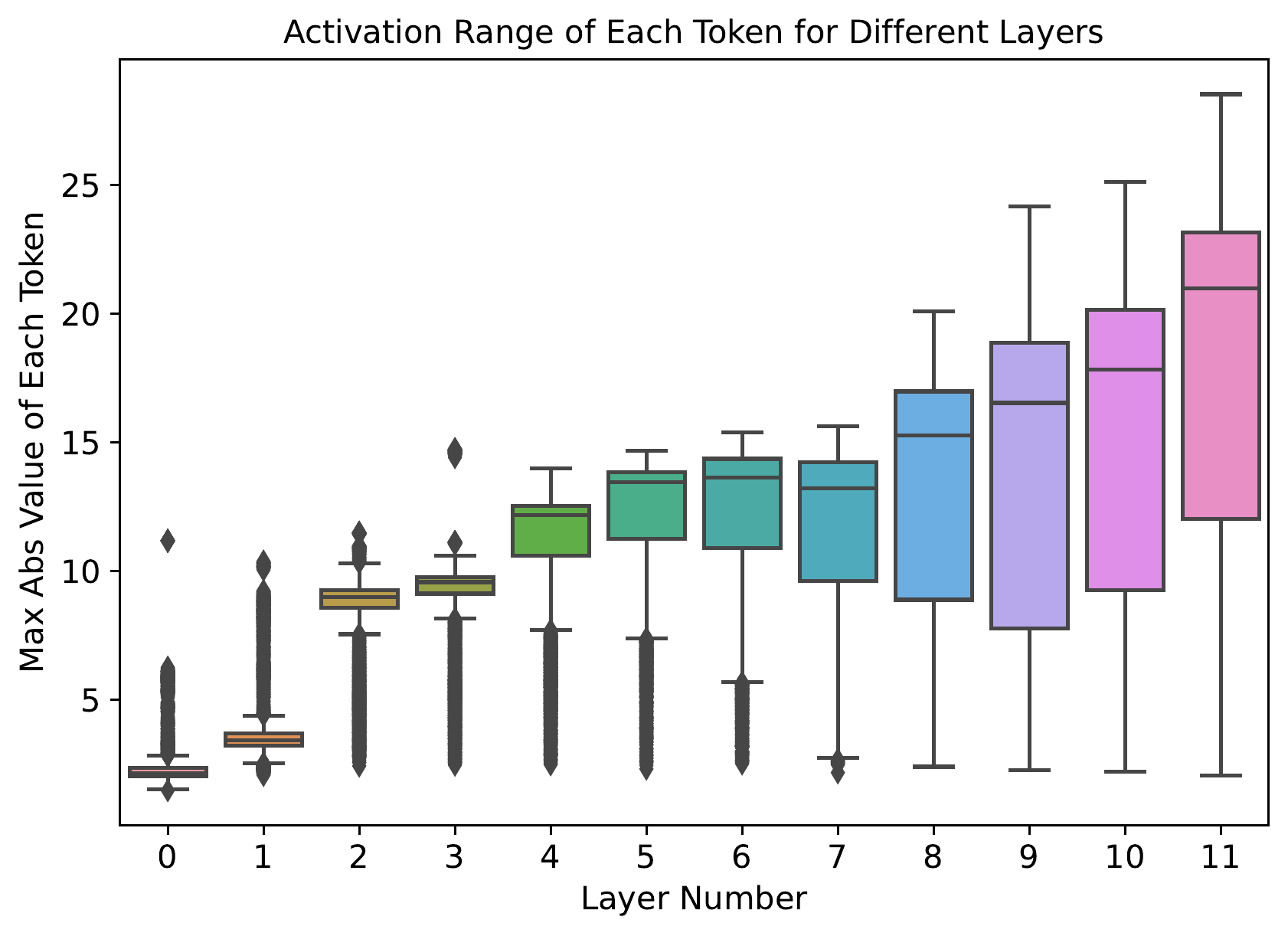}
\includegraphics[width=0.49\linewidth]{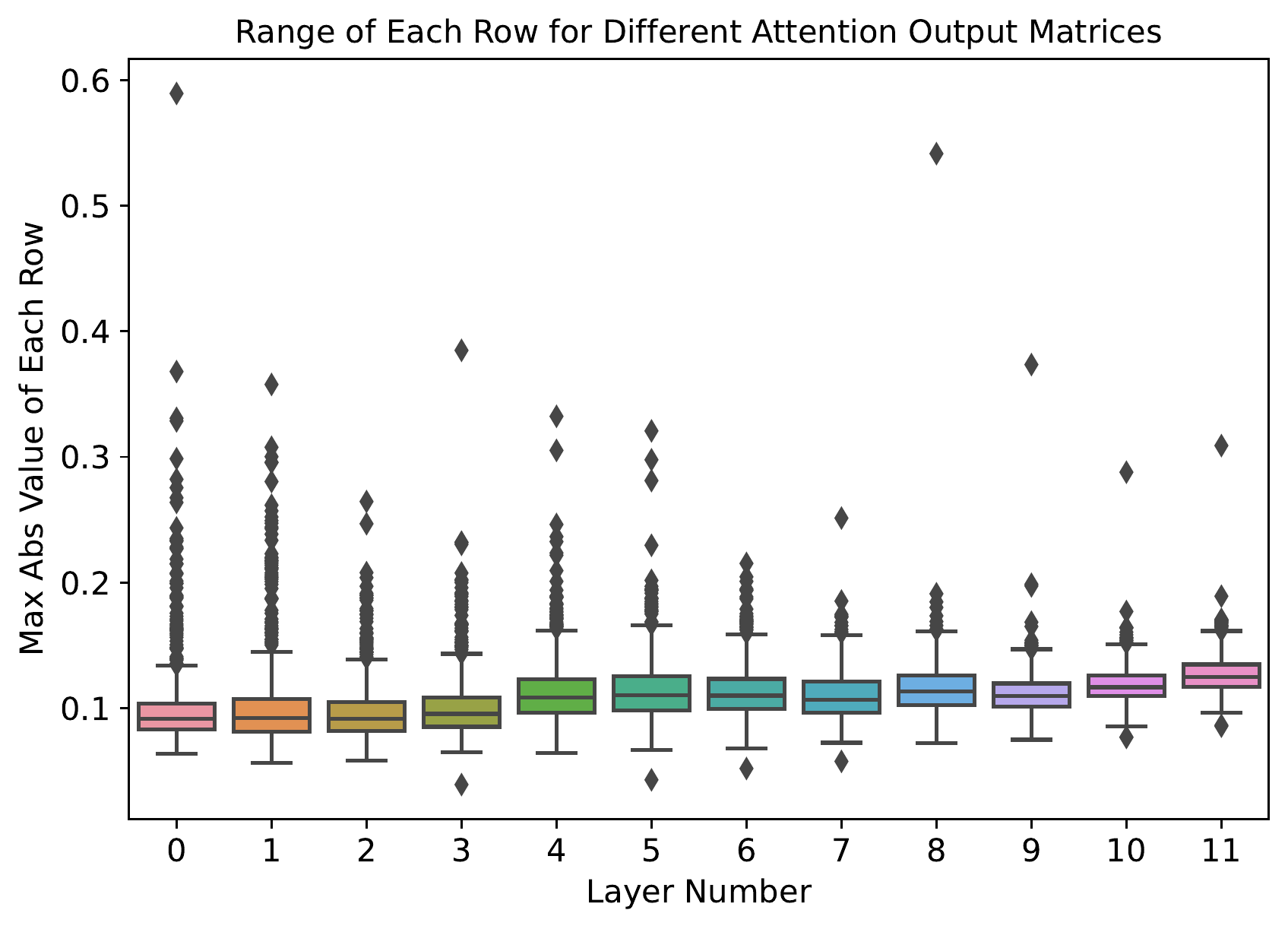}
\caption{
The activation range of different layers (left) and the row-wise weight range of the attention output matrix ($\mW_{o}$) of different layers (right). 
The results are based on the \bertbase trained on MNLI dataset. 
Please see~\fref{fig:activation_weight_range_gpt} for the results of \gpttf.
}
\label{fig:activation_weight_range}
\end{figure}

\section{Details about System Optimization}
\label{sec:details_about_system_optimization}
By having the weight and activation quantization, we can use the GeMM schedule that exploits the INT8 Tensor-core units which
provide 2x/4x more compute efficiency compared to the FP16/FP32 Tensor cores. 
For this purpose, we adapt the CUTLASS library to produce multiple schedules based on the input sizes we are considering in our application, such as the batch size, sequence length, and the Transformer hidden dimension. 
To achieve the best latency, we also develop our own efficient parallel implementation of the quantization operator on GPU.
During the inference run-time, based on the total batch size ($batch \times seq_len$), we choose the schedule that results in the lowest possible padding when performing the Tensor-core matrix-multiplication operations. 

To find the best schedule for the GeMM operation, we use the CUTLASS profiler tool that explores the tiling dimensions on
the thread-blocks, WARPs, and WMMA (Tensor cores), as the three compute hierarchies available within the Ampere GPU architecture.
Then, we find the best schedule by sorting the tile-based schedule based on either peak throughput achieved on the large-batch case, or
the maximum memory bandwidth taken from the main memory when the batch size is small.

However, there are still several challenges we need to address which are discussed below.

\paragraph{Operation Fusion for Token-wise Activation Quantization.} 
One of the main challenges of our quantization scheme is how to efficiently quantize hidden states before the GeMM operation. 
In order to remove the overhead, we fuse the activation quantization with its associated element-wise and/or reduction-based operations such as bias-addition, GELY, and LayerNorm.
This is due to the fact that
each SM takes care of one row (token) of the activation and therefore, we can reuse the computation from the thread registers
and compute the quantization scale, avoiding the data movement between GPU kernels and main memory. 
Moreover, by converting data from FP16 to INT8, we can utilize the memory bandwidth twice, further improving the inference latency and throughput.

\paragraph{Dequantization Associated with GeMM Schedule}
To utilize the output of integer output from GeMM operator in the following operators, one important step is to dequantize the output by using the scaling factor of the weight and activations. 
This dequantization step generally introduces extra overhead for quantized network inference due to the data movement. 
As such, we add a custom epilogue, which converts the final accumulated result (from INT32 format) of each row and column of the output to the 
real value (in FP16 format), using corresponding floating-point quantization scales computed from weight and activation group-wise quantization.
By fusing the dequantization with GeMM schedule, we ensure that there is no overhead exposed by using the INT8 operations while producing the
FP16 results that are used in the following operation.

Furthermore, to effectively combine dequantization with the GeMM operation, 
we read the two groups of quantization scales for the activation and weight matrices in advance prior to completion of the multiplication of the output matrix. 
Doing so, we overlap the reading of the extra quantization parameters with the 
GeMM computation and the GeMM-plus-dequantization can seamlessly work together without stalling the inference pipeline.

\paragraph{Cuda Graph Enhanced Small Model Inference.}
As the execution time for specific kernels reduce by optimizing the throughput using the INT8 inference pipeline, the overhead of launching
the GPU kernels and the CPU-to-GPU communication become a major bottleneck mostly on small-scale models. 
To address this issue, we add the CUDA-Graph support to our inference pipeline that reduces the CPU overhead, by storing the trace of the kernels launched during the inference 
forward computation, and creating the computation-graph to be reused in the next call to the inference pipeline. 
Thus, after storing the graph for 
the first time, we can replay the graph for the following requests, which substantially improves the performance especially on small models, such 
as \bertbase.
For a fair comparison, we also enable Cuda Graph for FP16 baseline.

\section{Tuned Results on BERT}
\label{sec:tuning_results_on_bert}
As mentioned in the main text and~\appref{sec:main_result_training_details}, we use the same set of hyperparameters for BERT. 
However, tuning them can significantly boost the performance for \OURS. 
Here, we tune two hyperparameters, i.e., the learning rate and the number of iterations in order to show the best possible performance of \OURS on both \bertbase and \bertlarge.
Particularly, we choose learning rate from the set \{1e-6, 2e-6, 5e-6, 1e-5\}, and choose number of iterations from the set \{0, 50, 100, 200, 400, 800, 1600\}. 
Thanks to the lightweight of \lwd, the total tuning time for \bertbase (including all data loading time, evaluation time, tokenization time, all three quantization schemes, etc) is around 4.5 hours on 8 40G-A100 GPUs (i.e., 36 GPU hours), and the tuning time for \bertlarge is around 16 hours on 8 40G-A100 GPUs (i.e., 128 GPU hours). 

We summarize the best results in the~\tref{tab:bert_base_tuned_result} and~\ref{tab:bert_large_tuned_result}. 

\begin{table}[t]
\caption{
Result of \bertbase on the development set of GLUE benchmark (except WNLI). 
Here WxAy means x-bit for weight quantization and y-bit for activation quantization.
Particularly, for W4/8, we quantize the \mhsa's weight to INT8 and \ffc's weight to INT4. 
Please see~\appref{sec:accuracy_reported_for_bert_on_glue} for the reported metrics.
}\centering
\label{tab:bert_base_tuned_result}
\begin{adjustbox}{width=0.99\linewidth}
\centering
\begin{tabular}{lcccccccccccccc }
\toprule
Precision (Method)   & CoLA  & MNLI-m & MNLI-mm & MRPC & QNLI & QQP & RTE & SST-2 & STS-B    & Ave.\\
\midrule
W32A32 (Baseline)   & 59.72 & 84.94 & 85.06 & 86.27/90.57 & 92.15 & 91.51/88.56 & 72.20 & 93.23 & 90.06/89.59 & 83.95\\
\midrule
W8A8 (\OURSlwd No Tuning)       & 59.59 & 84.83 & 85.13 & 86.03/90.39 & 91.98 & 91.45/88.46 & 71.12 & 93.12 & 90.09/89.62 & 83.75 \\
W8A8 (\OURSlwd Tuned) & 60.90 & 84.95 & 85.10 & 86.27/90.60 & 92.07 & 91.47/88.47 & 71.84 & 93.46 & 90.09/89.62 & 84.07 \\
\midrule
W4/8A32 (\OURSlwd No Tuning)     & 58.50 & 83.16 & 83.69 & 84.80/89.31 & 90.83 & 88.94/84.12 & 70.04 & 92.78 & 88.49/88.67 & 82.35 \\
W4/8A32 (\OURSlwd Tuned) & 60.04 & 83.64 & 84.31 & 85.78/89.53 & 91.01 & 90.66/87.26 & 71.84 & 93.12 & 88.68/88.79 & 83.26 \\
\midrule
W4/8A8 (\OURSlwd  No Tuning)      & 58.80 & 83.09 & 83.65 & 85.78/89.90 & 90.76 & 89.32/84.85 & 71.84 & 93.00 & 88.16/88.55 & 82.71 \\
W4/8A8 (\OURSlwd  Tuned)          & 60.30 & 83.47 & 84.03 & 85.78/89.90 & 90.87 & 90.77/87.38 & 71.84 & 93.00 & 88.38/88.70 & 83.22\\
\bottomrule
\end{tabular}
\end{adjustbox}
\end{table}

\begin{table}[t]
\caption{
Result of \bertlarge on the development set of GLUE benchmark (except WNLI). 
Here WxAy means x-bit for weight quantization and y-bit for activation quantization.
Particularly, for W4/8, we quantize the \mhsa's weight to INT8 and \ffc's weight to INT4. 
Please see~\appref{sec:accuracy_reported_for_bert_on_glue} for the reported metrics.
}
\centering
\label{tab:bert_large_tuned_result}
\begin{adjustbox}{width=0.99\linewidth}
\centering
\begin{tabular}{lcccccccccccccc }
\toprule
Precision (Method)   & CoLA  & MNLI-m & MNLI-mm & MRPC & QNLI & QQP & RTE & SST-2 & STS-B    & Ave. \\
\midrule
W32A32 (Baseline)   & 63.35 & 86.65 & 85.91 & 87.99/91.62 & 92.24 & 91.08/88.08 & 74.01 & 93.46 & 90.34/90.11 & 85.03  \\
\midrule
W8A8 (\OURSlwd No Tuning)       & 63.38 & 86.52 & 85.64 & 87.75/91.50 & 92.31 & 91.09/88.05 & 72.56 & 93.35 & 90.45/90.19 & 84.81 \\
W8A8 (\OURSlwd Tuned)           & 64.36 & 86.64 & 85.74 & 88.48/91.97 & 92.49 & 91.15/88.13 & 74.73 & 93.58 & 90.45/90.19 & 85.30 \\
\midrule
W4/8A32 (\OURSlwd No Tuning)     & 63.72 & 84.90 & 84.81 & 87.99/91.39 & 91.45 & 90.34/86.92 & 51.62 & 92.43 & 89.46/89.29 & 81.85 \\
W4/8A32 (\OURSlwd Tuned)         & 64.06 & 85.02 & 84.98 & 88.73/91.99 & 91.82 & 90.45/87.12 & 52.35 & 92.78 & 89.72/89.44 & 82.19 \\
\midrule
W4/8A8 (\OURSlwd No Tuning)      & 63.51 & 84.70 & 84.71 & 88.73/91.99 & 91.73 & 90.25/86.74 & 49.82 & 92.09 & 89.34/89.08 & 81.62 \\
W4/8A8 (\OURSlwd Tuned)          & 63.60 & 84.77 & 84.90 & 88.97/92.15 & 91.87 & 90.37/86.99 & 50.54 & 92.55 & 89.57/89.38 & 81.88 \\
\bottomrule
\end{tabular}
\end{adjustbox}
\end{table}

\section{\qat on \bertlarge}
\label{sec:qat_for_bearlarge}
We use four different learning rates for \qat on \bertlarge, \{5e-6, 1e-5, 2e-5, 5e-5\}. 
The final results we reported in the paper are chosen from the best single run among those four different learning rates. 
However, even with such tuning, we are not able to get good performance for \bertlarge on RTE.

Also, note that the time cost we used in the main text is based on a single run. 
if we consider the tuning cost, the total time will be $4\times 7181$s

\section{Limitations and Future Work}
\label{sec:limitations_and_future_work}
We believe it is critical for every work to clearly state its limitations, especially in this area. 
One limitation is that in this work we only focused on natural language models, but it would
be interesting to see how \OURS would perform for computer vision models. 
We leave this as a future work. 

Another limitation is that we can only verify the scalability of \OURS up to 20B scale models. 
If there are new releases of larger open-sourced models, it would be great to test \OURS on those larger models as well. 

Third, in this paper, we found out that the activation input of self-attention is more sensitive for quantization for the extra-large model (\gptneox). 
However, we are unable to verify this on other extra-large models due to the lack of open-sourced models.

\section{Full Zero-shot Evaluation of \gpt-style Models}

We includes all zero-shot evaluation results in this section for all \gpt-style models, inlcuding \gptneox. 

\begin{table}[t]
\caption{
The full results of \gpttf.
}\centering
\label{tab:gpt3_350_full_table}
\begin{adjustbox}{width=0.99\linewidth}
\centering
\begin{tabular}{lcccccccccccccccccccccccccccccccccccc}
\toprule
\multirow{2}{*}{Tasks}   & Baseline  & \multicolumn{4}{c}{PTQ} &  & \multicolumn{3}{c}{\OURS}  &  & \multicolumn{2}{c}{\OURSlwd}    \\
\cline{3-6}\cline{8-10} \cline{12-13}
                  & W32A32  & W8A32  & W32A8  & W8A8 & W4/8A32     & &W8A8 & W4/8A32 & W4/8A8   & & W4/8A32 & W4/8A8\\
HellaSwag         & 38.6    & 38.1	  & 37.6	& 36.8	& 26.5    & & 38.4	& 30.4	& 30.5  & & 35.3	& 35.3\\
LAMBADA           & 49.3    & 49.3	  & 44.7	& 42.9	& 0       & & 51.0	& 10.1	& 10.5  & & 39.8	& 37.4\\
TriviaQA          & 3.00    & 2.67	  & 2.70 	& 2.32	& 0       & & 2.86	& 0.159	& 0.194 & & 1.043	& 0.23\\
WebQs             & 1.43    & 0.935	  & 1.23	& 0.689	& 0       & & 1.378	& 0.246	& 0.394 & & 0.591	& 0.049\\
Winogrande        & 53.2    & 52.1	  & 52.1	& 52.1	& 47.8    & & 51.4	& 52.6	& 50.7  & & 51.6	& 51.8\\
PIQA              & 66.3    & 66.1	  & 64.8	& 64.1	& 51.4    & & 66.5	& 58.5	& 57.7  & & 63.8	& 61.8\\
ARC (Challenge)   & 24.2    & 24.0    & 24.0    & 24.1	& 27.0    & & 24.5	& 22.0	& 21.8  & & 21.8	& 23.6\\
ARC (Easy)        & 45.5    & 44.7	  & 44.2	& 43.9	& 25.1    & & 44.5	& 37.6	& 37.5  & & 40.5	& 40.5\\
ANLI R1           & 31.1    & 30.0    & 31.3	& 33.2	& 33.4    & & 31.1	& 32.8	& 32.7  & & 32.4	& 33.8\\
ANLI R2           & 34.3    & 36.0    & 36.5	& 35.9	& 33.4    & & 34.3	& 34.7	& 34.2  & & 34.1	& 33.5\\
ANLI R3           & 34.1    & 34.0    & 33.0    & 37.2	& 33.5    & & 33.4	& 34.9	& 34.5  & & 33.1	& 33.4\\
OpenBookQA        & 29.4    & 29.6	  & 28.2	& 28.0  & 30.2    & & 29.2	& 27.2	& 28.0  & & 29.4	& 28.2\\
RACE-h            & 32.4    & 31.3	  & 30.3	& 30.7	& 22.4    & & 32.2	& 25.7	& 26.4  & & 29.5	& 29.7\\
BoolQ             & 60.3    & 60.2	  & 57.0    & 56.9	& 37.8    & & 60.2	& 60.1	& 59.4  & & 61.9	& 61.9\\
Copa              & 69.0    & 67.0    & 71.0    & 73.0  & 48.0    & & 69.0	& 63.0	& 64.0  & & 68.0	& 66.0\\
RTE               & 53.8    & 54.2	  & 52.7	& 53.1	& 52.7    & & 53.4	& 52.0	& 52.7  & & 53.1	& 53.1\\
WSC               & 36.5    & 36.5	  & 36.5	& 35.6	& 63.5    & & 36.5	& 36.5	& 36.5  & & 36.5	& 36.5\\
MultiRC           & 0.839   & 0.839	  & 0.839	& 0.944	& 0.315   & & 0.839	& 1.889	& 1.889 & & 0.839	& 0.839\\
ReCoRD            & 75.1    & 74.8	  & 69.2	& 67.5	& 16.1    & & 74.9	& 56.5	& 55.3  & & 70.1	& 68.5\\
\wikitext      & 21.52   & 22.09	  & 24.56	& 26.20	& 1.76e5  & & 21.68	& 88.64	& 92.10 & & 30.56	& 31.13\\
Average Acc       & 38.86   & 38.54	  & 37.78	& 37.84	& 28.9    & & 38.71	& 33.52	& 33.42 & & 37.02	& 36.64\\
\bottomrule
\end{tabular}
\end{adjustbox}
\end{table}

\begin{table}[t]
\caption{
The full results of \gptot.
}\centering
\label{tab:gpt3_13_full_table}
\begin{adjustbox}{width=0.89\linewidth}
\centering
\begin{tabular}{lcccccccccccccccccccccccccccccccccccc}
\toprule
\multirow{2}{*}{Tasks}   & Baseline  & \multicolumn{2}{c}{PTQ} &  & \multicolumn{3}{c}{\OURS}  &  & \multicolumn{2}{c}{\OURSlwd}    \\
\cline{3-4}\cline{6-8} \cline{10-11}
                  & W32A32  & W8A8 & W4/8A32     & &W8A8 & W4/8A32 & W4/8A8   & & W4/8A32 & W4/8A8\\
HellaSwag         & 51.4    & 47.0	& 26.1    & & 50.8	& 43.7	& 43.2  & & 48.5	& 46.7 \\
LAMBADA           & 61.3    & 54.8	& 0       & & 62.6	& 43.9	& 46.8  & & 59.4	& 48.7 \\
TriviaQA          & 7.37    & 4.43	& 0       & & 6.67	& 2.36	& 2.09  & & 4.28	& 2.99 \\
WebQs             & 2.90    & 1.476 & 0       & & 2.07	& 1.132	& 1.28  & & 1.673	& 1.083 \\
Winogrande        & 57.1    & 55.7	& 50.1    & & 57.1	& 54.6	& 54.3  & & 55.3	& 53.8 \\
PIQA              & 71.4    & 67.7	& 50.4    & & 70.7	& 66.5	& 66.4  & & 69.5	& 68.1 \\
ARC (Challenge)   & 27.2    & 27.1	& 26.5    & & 26.8	& 25.7	& 25.3  & & 27.8	& 26.5 \\
ARC (Easy)        & 54.5    & 49.7	& 26.0    & & 53.8	& 48.0	& 47.0  & & 52.2	& 50.3 \\
ANLI R1           & 32.0    & 33.1	& 33.0    & & 33.4	& 33.8	& 33.6  & & 34.2	& 33.8 \\
ANLI R2           & 32.0    & 32.9	& 33.3    & & 33.9	& 33.0	& 33.0  & & 33.8	& 32.8 \\
ANLI R3           & 33.8    & 33.5	& 32.3    & & 34.8	& 33.6	& 33.5  & & 33.7	& 33.0 \\
OpenBookQA        & 33.6    & 32.6  & 27.0    & & 33.4	& 30.0	& 28.8  & & 31.6	& 29.0 \\
RACE-h            & 33.6    & 32.6	& 22.4    & & 32.7	& 30.9	& 29.9  & & 32.7	& 33.2 \\
BoolQ             & 62.4    & 59.2	& 37.8    & & 61.3	& 60.3	& 59.8  & & 61.7	& 61.3 \\
Copa              & 70.0    & 70.0  & 55.0    & & 72.0	& 73.0	& 74.0  & & 72.0	& 70.0 \\
RTE               & 53.1    & 54.5	& 50.9    & & 52.7	& 52.7	& 52.7  & & 52.7	& 52.0 \\
WSC               & 37.5    & 36.5	& 63.5    & & 36.5	& 36.5	& 36.5  & & 36.5	& 36.5 \\
MultiRC           & 1.05    & 0.839 & 0.315   & & 0.839	& 1.259	& 1.154 & & 0.839	& 0.839 \\
ReCoRD            & 82.6    & 75.7	& 15.8    & & 80.9	& 77.3	& 76.2  & & 79.7	& 77.4 \\
\wikitext      & 15.3    & 18.85 & 1.35e5  & & 15.69	& 21.9	& 24.09 & & 17.56	& 18.18 \\
Average Acc       & 42.36   & 40.49 & 28.97   & & 42.26	& 39.38	& 39.24 & & 41.48	& 39.90 \\
\bottomrule
\end{tabular}
\end{adjustbox}
\end{table}

\begin{table}[t]
\caption{
The full results of W4/8A8 \gpttf using different data resources.
}\centering
\label{tab:gpt3_350_data_ablation_study_full_table}
\begin{adjustbox}{width=0.6\linewidth}
\centering
\begin{tabular}{lcccccccccccccccccccccccccccccccccccc}
\toprule
Tasks   & Random Data  & Wikipedia  & Original Training Data \\
                  
HellaSwag        & 33.9  & 35.5   & 35.3  \\
LAMBADA          & 26.1  & 33.9   & 37.4  \\
TriviaQA         & 0.088 & 0.972  & 0.23  \\
WebQs            & 0.049 & 0.344  & 0.049 \\ 
Winogrande       & 50.3  & 52.4   & 51.8  \\
PIQA             & 59.3  & 62.4   & 61.8  \\
ARC (Challenge)  & 22.6  & 23.3   & 23.6  \\
ARC (Easy)       & 38.3  & 40.0   & 40.5  \\
ANLI R1          & 33.0  & 32.0   & 33.8  \\
ANLI R2          & 34.3  & 34.7   & 33.5  \\
ANLI R3          & 33.4  & 32.9   & 33.4  \\
OpenBookQA       & 29.2  & 28.0   & 28.2  \\
RACE-h           & 27.8  & 29.1   & 29.7  \\
BoolQ            & 47.8  & 52.6   & 61.9  \\
Copa             & 65.0  & 69.0   & 66.0  \\
RTE              & 50.5  & 52.7   & 53.1  \\
WSC              & 36.5  & 36.5   & 36.5  \\
MultiRC          & 1.574 & 1.154  & 0.839 \\ 
ReCoRD           & 64.9  & 69.5   & 68.5  \\
\wikitext     & 40.63 & 30.36  & 31.13 \\ 
Average Acc      & 34.45 & 36.16  & 36.64 \\ 
\bottomrule
\end{tabular}
\end{adjustbox}
\end{table}

\begin{table}[t]
\caption{
The full results of \gptneox.
}\centering
\label{tab:gptneox_full_table}
\begin{adjustbox}{width=0.4\linewidth}
\centering
\begin{tabular}{lcccccccccccccccccccccccccccccccccccc}
\toprule
Tasks   & W16A16  & W8A8/16 \\
                  
HellaSwag        & 71.4  & 71.2 \\
LAMBADA          & 71.7  & 71.9 \\
TriviaQA         & 25.8  & 25.9 \\
WebQs            & 6.3   & 6.64 \\
Winogrande       & 66.0  & 65.7 \\
PIQA             & 77.7  & 78.3 \\
ARC (Challenge)  & 41.0  & 42.2 \\
ARC (Easy)       & 68.5  & 68.8 \\
ANLI R1          & 33.1  & 33.9 \\
ANLI R2          & 33.4  & 34.4 \\
ANLI R3          & 35.1  & 35.4 \\
OpenBookQA       & 39.8  & 38.8 \\
RACE-h           & 38.5  & 37.6 \\
BoolQ            & 69.4  & 69.9 \\
Copa             & 84.0  & 85.0 \\
RTE              & 54.9  & 54.9 \\
WSC              & 50.0  & 44.2 \\
MultiRC          & 3.57  & 4.41 \\
ReCoRD           & 88.3  & 88.0 \\
Average Acc      & 50.45 & 50.38 \\
\bottomrule
\end{tabular}
\end{adjustbox}
\end{table}

\end{document}